\documentclass[10pt,letterpaper,twocolumn]{article}
\usepackage[T1]{fontenc}
\usepackage[utf8]{inputenc}
\usepackage{lmodern}
\usepackage[margin=0.8in,columnsep=0.26in]{geometry}
\usepackage{microtype}
\usepackage{balance}
\usepackage[round]{natbib}
\usepackage{amsmath,amssymb,amsthm,mathtools, bbm}
\usepackage{algorithm,algpseudocode}
\usepackage{booktabs,array,enumitem,graphicx,subcaption,xcolor,xspace}
\usepackage{hyperref}
\usepackage{cleveref}
\usepackage{authblk}
\hypersetup{hidelinks,pdftitle={Calibrating Ambiguity Set via Diagnostic Transport for Distributionally Robust Optimization},pdfauthor={Wenbin Zhou, Elizabeth Cucuzzella, Shixiang Zhu}}
\newcommand{\R}{\mathbb R}

\renewcommand{\P}{\mathbb P}
\newcommand{\Q}{\mathbb Q}
\newcommand{\E}{\mathbb E}
\newcommand{\1}{\mathbbm 1}

\newcommand{\cZ}{\mathcal Z}
\newcommand{\cX}{\mathcal X}
\newcommand{\cY}{\mathcal Y}
\newcommand{\cG}{\mathcal G}
\newcommand{\cD}{\mathcal D}
\newcommand{\cU}{\mathcal U}

\newcommand{\W}{\mathsf W_1}
\renewcommand{\d}{\mathrm d}
\newcommand{\dd}{\mathrm d}
\newcommand{\Pb}{\P_{b,x}}
\newcommand{\Ps}{\P_x^\star}
\newcommand{\Ph}{\widehat\P_x}
\newcommand{\Phm}{\widehat\P_x^{(m)}}
\newcommand{\Pbm}{\P_{b,x}^{(m)}}
\newcommand{\Gh}{\widehat G_x}
\newcommand{\Gs}{G_x^\star}
\newcommand{\Th}{\widehat T_x}
\newcommand{\D}{\Delta_{\cY}}

\DeclareMathOperator*{\argmin}{arg\,min}

\newcommand{\abs}[1]{\left|#1\right|}
\theoremstyle{plain}
\newtheorem{theorem}{Theorem}
\newtheorem{proposition}{Proposition}
\newtheorem{lemma}{Lemma}

\theoremstyle{definition}
\newtheorem{assumption}{Assumption}

\theoremstyle{remark}

\crefname{assumption}{Assumption}{Assumptions}
\Crefname{assumption}{Assumption}{Assumptions}
\title{Calibrating Ambiguity Set via Diagnostic Transport for \\ Distributionally Robust Optimization}
\author[1, 2]{Wenbin Zhou}
\author[3]{Elizabeth Cucuzzella}
\author[1]{Shixiang Zhu}
\affil[1]{Heinz College of Information Systems and Public Policy, Carnegie Mellon University}
\affil[2]{Machine Learning Department, Carnegie Mellon University}
\affil[3]{Department of Statistics and Data Science, Carnegie Mellon University}

\date{}
\begin{document}
\maketitle
\pagestyle{plain}
\raggedbottom

\begin{abstract}
Distributionally robust optimization (DRO) protects decisions against distributional uncertainty by optimizing over an ambiguity set, but poorly aligned set geometry can require large radii and yield overly conservative decisions. We introduce diagnostic-transport DRO (DT-DRO), which uses held-out calibration data to adapt the ambiguity-set geometry to observed predictive errors. DT-DRO uses the conditional probability integral transform cumulative distribution function to diagnose systematic probability misallocation and translates this information into an outcome-level transport that jointly adjusts the ambiguity-set center and ground cost. The resulting formulation admits a computationally tractable dual reformulation. Theoretically, we derive valid ambiguity radii and decision-risk guarantees that tighten as estimation and approximation errors vanish, and show that DT-DRO can eliminate the nonvanishing robustness floor caused by model misspecification. Synthetic experiments and a power-outage application demonstrate improved decision quality, particularly under structural and tail misspecification.
\end{abstract}

\section{Introduction}
\label{sec:introduction}

Operational decisions often rely on contextual information: an electricity operator procures reserves from net-load forecasts \citep{hobbs2022probabilistic}, a retailer sets inventory from demand forecasts \citep{prak2019forecasting}, and an emergency manager allocates supplies using hurricane forecasts \citep{davis2013inventory}. In these cases, the decision maker must account not only for the predicted outcome, but also for uncertainty and potential misspecification in the predictive distribution on which the decision is based.

Distributionally robust optimization (DRO) addresses this problem by optimizing decisions against distributions in an ambiguity set, with finite-sample guarantees and tractable reformulations under suitable conditions \citep{mohajerin2018data}. In predictive settings, this set is often configured as a Wasserstein or optimal-transport ball around a base predictive distribution \citep{gao2023distributionally, blanchet2022optimal}. The ball is typically sized to contain the true predictive distribution with high probability, after which the decision is chosen to hedge against the worst-case distribution within that set.

A well-shaped ambiguity-set geometry that aligns with the structure revealed by the data is critical for obtaining informative decisions. If the geometry is poorly aligned with the underlying distributional uncertainty, a larger radius may be required to ensure that the true distribution is covered, thereby expanding the ambiguity set beyond what is necessary. This issue often arises when implementing standard DRO, where the geometry is often specified a priori through a fixed ground cost without using the observed data to adapt its shape. The resulting ambiguity set may therefore include many implausible adverse distributions, forcing the decision maker to hedge against uncertainty that is not supported by the data and leading to overly conservative decisions with poor realized performance.

\begin{figure}[t]
 \centering
 \includegraphics[width=\linewidth]{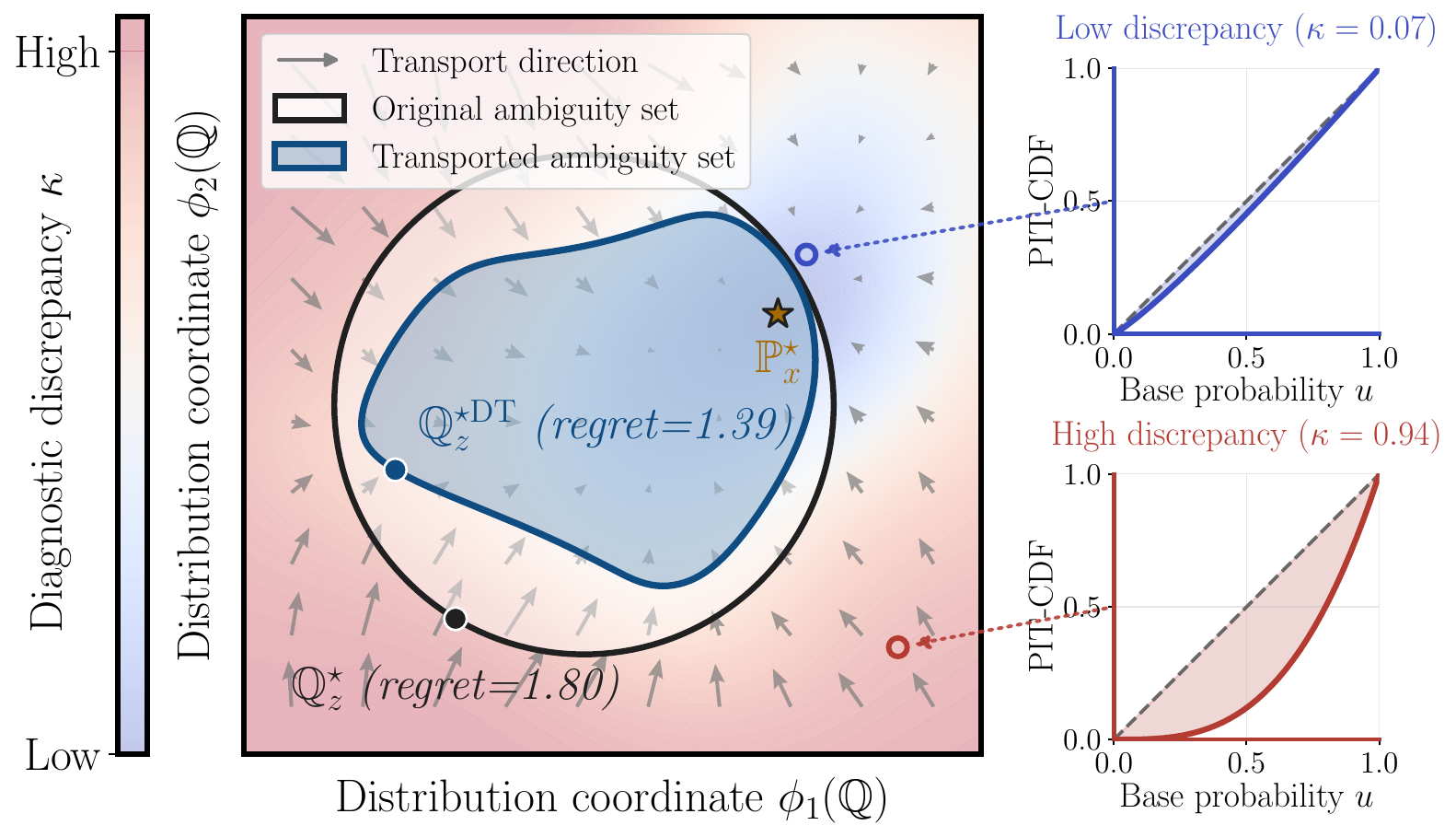}
 \caption{
A synthetic illustration of diagnostic transport in DT-DRO. The heatmap visualizes the diagnostic discrepancy $\kappa$, scaling with the shaded area between the PIT-CDF and the identity line. Gray arrows indicate the local transport directions in schematic distribution-space coordinates, illustrating how diagnostic information reshapes the ambiguity set. The transported ambiguity set (blue region) contracts toward the true distribution $\Ps$, and the corresponding worst-case distribution $\Q_z^{\star,\rm DT}$ yields lower regret than the worst-case distribution $\Q_z^\star$ obtained from the original ambiguity set. }
 \label{fig:transport}
\end{figure}

To tackle this issue, we introduce \emph{diagnostic-transport DRO} (DT-DRO), which uses data unseen during training to calibrate the transport-geometry that defines the ambiguity set. The key observation is that the base predictive model already provides a natural coordinate system for diagnosing how the ambiguity set should be reshaped: by examining the conditional probability integral transform, we can identify where the model systematically over- or under-allocates probability mass relative to observed outcomes. These discrepancies are summarized by the conditional probability integral transform cumulative distribution function (PIT-CDF), which reveals both the magnitude and direction of local distributional misalignment. DT-DRO converts this diagnostic information into an outcome-level transport that shifts the ambiguity set toward contextual calibration. See \Cref{fig:transport} for an illustration.

Importantly, the transported ambiguity set need not be constructed explicitly. We show that DT-DRO retains a tractable dual reformulation because the transported set is equivalent to an ambiguity set with a transported center and an induced ground cost. Moreover, computing the transport itself reduces to a statistical estimation problem: for scalar outcomes, the optimal map admits a closed-form representation determined by the conditional PIT-CDF of the base model \citep{dey2025ladar,cucuzzella2026trustworthy}. Theoretically, we characterize a valid radius and decision-risk certificate gap that both vanish as diagnostic-estimation and scenario-approximation errors decrease, highlighting the statistical benefit of DT-DRO over model-refitting strategies when the predictive model remains systematically misspecified.

Experiments on synthetic benchmarks and a real-world power-outage resilience application demonstrate that DT-DRO consistently improves decision quality under predictive misspecification. Across settings with location shift, heteroskedasticity, tail misspecification, and rare-event structure, the diagnostic wrapper yields lower regret than fixed-base and retrained Wasserstein-DRO baselines, with the largest gains arising when misspecification cannot be removed by refitting alone. The outage study further shows that DT-DRO maintains lower and more stable regret over time, including for severe upper-tail events, while improving upon center calibration alone through its transport-induced ambiguity geometry.

\paragraph{Related Work}
\label{sec:related}

Contextual optimization uses covariates to inform decisions under uncertainty \citep{bertsimas2020predictive}. Contextual DRO accounts for errors in conditional distribution estimation through probability trimmings and partial transport \citep{esteban2022trimmings}, regression residuals \citep{kannan2024residuals}, joint covariate--outcome transport \citep{nguyen2025robustifying}, and causal transport that preserves the nominal conditional information structure \citep{yang2026causal}. Data-driven transport-cost selection \citep{blanchet2019costselection} and geometry-calibrated robust objectives \citep{liu2024geometry} further adapt robustness to the structure of the data. Related conformal approaches attach score-calibrated uncertainty representations to fixed point predictors for robust optimization and robust satisficing \citep{zhao2026conformalrobustness}, or use nested conformal regions to construct ambiguity sets for latent instance-specific distributions \citep{zhang2026conformaldro}. DT-DRO instead starts from a fitted full predictive distribution and learns a common outcome transport that acts on every distribution in an ambiguity set. Therefore, our coverage relies on diagnostic-estimation control, rather than distribution-free conditional validity.

Decision-focused learning (DFL) takes a complementary approach by training predictive models against downstream decision loss rather than predictive accuracy alone. Task-based stochastic optimization learns probabilistic models through the ultimate optimization objective \citep{donti2017task}, while decision-loss surrogates \citep{elmachtoub2022smart} and differentiable relaxations \citep{wilder2019melding} extend this principle to predict-then-optimize, combinatorial problems, and generative model learning \citep{wang2026gendfl}. The same task-dependent principle motivates learning contextual uncertainty sets \citep{wang2023decisionfocused}, polyhedral conformal sets with independent recalibration \citep{chen2026polyhedral}, and end-to-end conformal uncertainty representations \citep{yeh2025endtoend}; loss-aware OT-DRO similarly learns ambiguity-set geometry through a bilevel decision objective \citep{ohnemus2026lossaware}. These approaches can improve performance for the chosen task, but they couple predictor or uncertainty-model fitting to downstream optimization. We adopt a different principle: the base predictive model remains fixed, and we learn a wrapper applied to the entire ambiguity set and solve the decision problem only after this map is fixed. It is particularly useful when task-specific retraining is undesirable, modification of the base model is not permitted, or repeated refitting within a misspecified model family would retain systematic error.

Finally, this work is related to probabilistic calibration, which corrects predicted probabilities based on observational data and is assessed via predictive distributions, scores, and quantiles \citep{dawid1984prequential,gneiting2007strictly,kuleshov2018accurate,song2019distribution}. While uniformity of the probabilistic integral transform (PIT) variable guarantees marginal calibration, it does not ensure correctness at an individual context, motivating local diagnostics \citep{zhao2021diagnostics}. LADaR learns a conditional probability--probability map that diagnoses and reshapes an initial predictive distribution \citep{dey2025ladar}; these diagnostic transport maps provide interpretable calibration \citep{cucuzzella2026trustworthy}. We adopt this predictive distribution calibration construction and contribute to understanding its action on an entire DRO ambiguity set, the uncertainty left by nonparametric estimation, and the consequences for robust decisions.

\section{Problem Setup}
\label{sec:setup}
We consider a decision maker who observes contextual information $x\in\cX$ and chooses a decision $z\in\cZ$ before an uncertain outcome $Y\in\cY$ is realized. Given a loss function $\ell(z,y)$, the decision maker seeks to minimize the worst-case expected loss over the plausible conditional distributions in an ambiguity set:
\begin{equation}
\label{eq:dro}
 \min_{z\in\cZ}\sup_{\Q\in\cU_x(\varepsilon)}\int\ell(z,y)\,\Q(\dd y).
\end{equation}
Here, $\cU_x(\varepsilon)$ contains the distributions of interest at context $x$, and $\varepsilon$ controls the size of the set. We assume that we have access to a data set of contexts and historical realized outcomes $\mathcal{D} = \{(x_i, y_i)\}_{i=1}^N$ to inform the decision-making process. A subset of samples in $\mathcal{D}$ might be used for training, while the remaining are reserved for calibration and evaluation.

In optimal-transport DRO \citep{blanchet2022optimal}, the ambiguity set is configured as a transport ball centered at a base predictive distribution $\Pb$:
\begin{equation}
\label{eq:base-ball}
 \cU_x(\varepsilon)=\left\{\Q:\mathsf D_c(\Q,\Pb)\le\varepsilon\right\},
\end{equation}
where the optimal transport (OT) distance is
\begin{equation}
\label{eq:w-dist}
 \mathsf D_c(\Q_1,\Q_2)
 =\inf_{\pi\in\Pi(\Q_1,\Q_2)}
 \int c(y,y')\,\pi(\dd y,\dd y').
\end{equation}
The function $c(y,y')$ is the cost of moving an outcome from $y$ to $y'$, and $\Pi(\Q_1,\Q_2)$ contains the couplings with marginals $\Q_1$ and $\Q_2$. The base model $\Pb$ represents the decision maker's estimate of the unknown true conditional distribution $\Ps$. For scalar outcomes with $c(y,y')=|y-y'|$, $\mathsf D_c(\Q_1,\Q_2)$ coincides with the type-1 Wasserstein distance, which we denote as $\W$.

The geometry of the ambiguity set is primarily determined by the transportation cost, which the decision maker often specifies in advance, for example as an $\ell_2$ distance. However, this choice is often suboptimal: the resulting geometry may not be flexible enough to capture the plausible uncertain distributions, requiring a large radius $\varepsilon$ and causing the decision maker to hedge against nonexistent uncertainty and select an overly conservative decision with a high realized loss.

\section{Proposed Framework}
\label{sec:method}
We address this challenge by proposing \emph{diagnostic-transport DRO} (DT-DRO), a framework that uses data unseen during training of the base model to calibrate the geometry of DRO ambiguity sets. The central idea is to define a diagnostic transport mapping \citep{zhao2021diagnostics,dey2025ladar,cucuzzella2026trustworthy} that acts as a wrapper over the entire ambiguity set to reshape the set. This construction has two key benefits: ($i$) it reduces over-conservatism in the resulting decision, improving its realized performance; and ($ii$) it preserves the ambiguity-set structure needed for tractable computational reformulations.

\subsection{Diagnostic-transport ambiguity set}

Our goal is to find such a transport mapping $T_x$ by solving a constrained optimal-transport problem.
\begin{equation}
\label{eq:center-ot}
\begin{aligned}
 \min_{T_x:\cY\to\cY}\quad
 &\E_{Y\sim\Pb}\!\left[|T_x(Y)-Y|^2\right]\\
 \text{s.t.}\quad
 &(T_x)_\#\Pb=\Ps,
\end{aligned}
\end{equation}
where $(T_x)_\#$ is the induced pushforward mapping:
\begin{equation}
\label{eq:tx-hash}
 (T_x)_\#\Q(A)=\Q(T_x^{-1}(A)).
\end{equation}
Intuitively, \eqref{eq:center-ot} aims to find the $T_x$ that shifts the outcome distributions toward the true conditional distribution while introducing the smallest possible average change to the outcomes.

Once \eqref{eq:center-ot} is solved, we apply $(T_x)_\#$ to every distribution in the ambiguity set, yielding
\begin{equation}
\label{eq:udt}
 \cU_x^{\rm DT}(\varepsilon)
 =
 \left\{(T_x)_\#\Q:\Q\in\cU_x(\varepsilon)\right\}.
\end{equation}
We then replace the original set $\mathcal{U}_x(\epsilon)$ in \eqref{eq:dro} with this transported set to define DT-DRO. Note that \eqref{eq:udt} need not be computed explicitly, as detailed next.

\subsection{Duality reformulation}
\label{sec:duality}

The transported ambiguity set defined in \eqref{eq:udt} retains an optimal-transport ball interpretation.

\begin{lemma}[Transported geometry]
\label{prop:ball}
Let $T_x$ be a measurable bijection with measurable inverse, $\mathcal{U}_x(\varepsilon)$ be defined in \eqref{eq:base-ball}. Set $\P_x^T=(T_x)_\#\Pb$ and
\begin{equation}
\label{eq:metric}
 c_{T_x}(y,y')=c(T_x^{-1}(y),T_x^{-1}(y')).
\end{equation}
Then \eqref{eq:udt} can be equivalently expressed as
\begin{equation}
\label{eq:ball}
 \cU_x^{\rm DT}(\varepsilon)
 =\{\Q:\mathsf D_{c_{T_x}}(\Q,\P_x^T)\le\varepsilon\}.
\end{equation}
\end{lemma}

\Cref{prop:ball} reveals that the effect of the transport can be decomposed into two parallel components: ($i$) shifting the base-model center through the pushforward mapping $(T_x)_\#$, and ($ii$) recalibrating the OT distance so that discrepancies between outcomes are measured on the transported scale induced by $T_x$.

This structure enables an explicit reformulation to be established following the standard duality derivation \citep{gao2023distributionally, blanchet2022optimal}:
\begin{equation}
    \label{eq:dual}
      \inf_{\substack{z\in\cZ \\ \lambda\ge0}}
     \left\{\lambda\varepsilon +
     \int\!\sup_{y\in\cY}
     \Big[\ell(z,y) - \lambda c_{T_x}(y,u)\Big] \P_x^T(\dd u)\right\}.
\end{equation}
In practice, one can approximate the integral using $m$ independently drawn scenarios $s_1,\ldots,s_m\sim\P_x^T$, and leverage several standard approaches, including discretization, local-reduction methods, and exchange or cutting-plane schemes \citep{hettich1993semi,xu2018distributionally,mohajerin2018data}, to solve the induced semi-infinite program.

In particular, under a convex piecewise-affine loss and a piecewise-affine transport map, \eqref{eq:dual} admits an exact linear-program (LP) reformulation, which enables efficient computation with standard LP solvers.

\begin{theorem}[Exact LP reduction]
\label{cor:lp-reduction}
Let $\cY=[a,b]$, $c(y,y')=|y-y'|$, and $\cZ$ be polyhedral. Suppose $T_x$ is continuous piecewise affine with knots $a=\tau_0<\cdots<\tau_K=b$, and the decision loss function is
\[
\ell(z,y)=\max_{r\in[R]}\{\alpha_r^\top z+\beta_r y+\gamma_r\}.
\]
For source scenarios $\nu_1,\ldots,\nu_m\sim\P_{b,x}$, define $ \mathcal K_i := \{\tau_0,\ldots,\tau_K,\nu_i\}. $ Then, the sample average approximation of \eqref{eq:dual} is a finite linear program
\begin{equation}
\label{eq:finite-lp}
\begin{aligned}
\min_{\substack{z\in\cZ,\lambda\ge0\\h\in\R^m}}
&\quad
\lambda\varepsilon+\frac{1}{m}\sum_{i=1}^m h_i\\
\mathrm{s.t.}
&\quad
h_i
\ge
\alpha_r^\top z
+\beta_r T_x(v)
+\gamma_r
-\lambda|v-\nu_i|,\\
&\quad
\forall v\in\mathcal K_i,\
\forall r\in[R],\
\forall i\in[m].
\end{aligned}
\end{equation}
\end{theorem}

\begin{algorithm}[t]
\caption{Diagnostic-transport DRO}
\label{alg:dtdro}
\begin{algorithmic}[1]
\Require Base CDF $F_b$, dataset $\cD$, context $x$, scenario count $m$, miscoverage rate $\alpha$.
\State Compute $u_i=F_b(y_i\mid x_i)$.
\State Fit a nonparametric increasing $\widehat G$ using \eqref{eq:loss}.
\State Form $\Th=F_b^{-1}\circ\Gh^{-1}\circ F_b$ at $x$.
\State Generate $m$ scenarios $s_i\sim\Pb$.
\State Select the radius according to \Cref{sec:radius}.
\State Solve \eqref{eq:dual} via sample average approximation.
\State \Return Decision $\widehat z$ and robust objective $\widehat v$.
\end{algorithmic}
\end{algorithm}

\subsection{Computational algorithm}
\label{sec:computation}
\label{sec:diagnostic}
\label{par:learning-diagnostic}

The algorithm can be described in two steps: ($i$) estimate the transport map $\smash{\widehat T_x}$ by solving \eqref{eq:center-ot}, and then ($ii$) substitute it into \eqref{eq:dual} to solve for the final decision. Under scalar outcomes, we show that $\smash{\widehat T_x}$ admits a closed-form solution, effectively simplifying the first-step into a pure statistical estimation problem.

\begin{lemma}[Diagnostic transport]
\label{prop:minimum-displacement}
Fix $x$ and suppose the base and true conditional CDFs are continuous and strictly increasing on $\cY=[a,b]$. Define the base-model conditional PIT and its CDF by
\begin{equation}
\label{eq:pit-map}
 U=F_b(Y\mid X),\quad
 G_x^\star(u)=\Pr(U\le u\mid X=x).
\end{equation}
Then an optimal solution for \eqref{eq:center-ot} is
\begin{equation}
\label{eq:tx-oracle}
 T_x^\star(y)
 =F_b^{-1}\!\left((G_x^\star)^{-1}(F_b(y\mid x))\mid x\right),
\end{equation}
where $G_x^\star:[0,1]\to[0,1]$ is the conditional PIT-CDF.
\end{lemma}

\Cref{prop:minimum-displacement} states that $T_x$ is a composition of three functions: $F_b(\cdot \mid x)$ first maps an outcome $y$ to its probability level under the base model, $(G_x^\star)^{-1}$ then adjusts this probability level according to the discrepancy revealed by the conditional PIT distribution, and $F_b^{-1}(\cdot\mid x)$ finally maps the corrected probability level back to the outcome space. The proof follows from the fact that $F^\star_x(y) = G^\star_x(F_b(y\mid x))$, which implies that if $F_b$ is perfectly conditionally calibrated, then the conditional PIT-CDF $G_x^\star$ is precisely the identity map on $[0,1]$. This property of $G_x^\star$ allows it to be used as a diagnostic tool for visualizing local distributional misalignment in the literature known as the conditional P-P plot \citep{zhao2021diagnostics}, motivating the term \emph{diagnostic transport} of $T_x$ in our framework.

\begin{figure}[t]
 \centering
 \includegraphics[width=\linewidth]{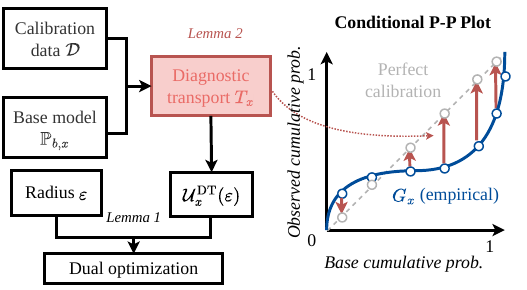}
 \caption{Overview of the DT-DRO procedure. The calibration data $\mathcal D$ and base predictive model $\P_{b,x}$ are used to estimate the conditional PIT-CDF $G_x$ and construct the diagnostic transport $T_x$. The conditional P--P plot visualizes the discrepancy between $G_x$ and perfect calibration, while the red arrows illustrate the calibration corrections induced by solving \eqref{eq:center-ot}. The learned transport reshapes the ambiguity set into $\mathcal U_x^{\rm DT}(\varepsilon)$; together with the calibrated radius $\varepsilon$, this transported set enters the dual optimization to produce the final robust decision.}
 \label{fig:pp-plot}
\end{figure}

To estimate $T_x$, it suffices to estimate $G_x$, since the base CDF $F_b$ and its inverse $F_b^{-1}$ are already known. By its definition in \eqref{eq:pit-map}, estimating $G_x$ can be formulated as a binary regression problem:
$$
G_x^\star(u)=\E[\1\{U\le u\}\mid X=x]
$$
This consists two steps: ($i$) compute $u_i=F_b(y_i\mid x_i)$ from the dataset $\mathcal D$ and ($ii$) fit the conditional CDF of $u_i$ by, for example, minimizing the integrated Brier loss \citep{brier1950}:
\begin{equation}
\label{eq:loss}
 \widehat G\in\argmin_{G\in\cG}
 \frac1n\sum_{i=1}^n\int_0^1
 \left[\1\{u_i\le u\}\!-\!G_{x_i}(u)\right]^2\dd u.
\end{equation}
The integral in \eqref{eq:loss} can be approximated by sampled thresholds, as in amortized local P--P estimation \citep{dey2025ladar}. Suitable model classes $\mathcal G$ include monotone neural networks, isotonic or spline models with context-dependent coefficients, and monotonic ensemble-based methods. In small-sample settings, one may instead estimate $\widehat G$ semiparametrically by specifying a parametric family for the transformation and modeling its parameters flexibly as functions of the context \citep{cucuzzella2026trustworthy}.

Finally, substituting the estimated diagnostic gives the estimated outcome-level transport map
\begin{equation}
\label{eq:tx-estimated}
 \Th(y)=F_b^{-1}\!\left(\Gh^{-1}(F_b(y\mid x))\mid x\right).
\end{equation}
The algorithmic procedure for solving DT-DRO is summarized as pseudocode in \Cref{alg:dtdro}, and the high-level pipeline is illustrated in \Cref{fig:pp-plot}.

\section{Theoretical Analysis}
\label{sec:theory}

In this section, we characterize the radius required to ensure distributional containment and derive the corresponding risk certificate and excess decision loss. We also examine the statistical benefit of diagnostic calibration relative to the refitting strategies commonly used in contextual DRO when additional target-population data become available \citep{kannan2024residuals,zhu2026decisiondependent}. Throughout, we work at a fixed context $x$, ground cost $c(y,y')=|y-y'|$, and conditional on the fitted base predictor. All detailed proof and derivations can be found in Appendix~\ref{app:proofs}.

\subsection{Valid radius selection}
\label{sec:radius}

We begin by imposing three assumptions on the support, monotonicity, and sampling scheme. Discussion of potential relaxations is deferred to Appendix~\ref{app:ass}.

\begin{assumption}[Regularity conditions]
    \label{ass:diagnostic}
$~$
    \begin{enumerate}
\item[($a$)] The outcome space is a bounded interval $\cY=[a,b]$, with $\D=b-a<\infty$. \item[($b$)] For each fixed context $x$, the base CDF $F_b(\cdot\mid x)$ and population diagnostic CDF $G_x^\star$ are continuous and strictly increasing, and the fitted diagnostic $\Gh$ is an increasing homeomorphism of $[0,1]$. \item[($c$)] The calibration pairs are i.i.d. and independent of the data used to fit the base model.
    \end{enumerate}
\end{assumption}

Additionally, we assume the existence of a constant $r_n(x,\delta)$ that is a high-probability bound characterizing the convergence of the estimated PIT-CDF to its population counterpart:
\begin{equation}
    \label{eq:maprate}
     \Pr\{\|\Gh-\Gs\|_\infty\le r_n(x,\delta)\}\ge1-\delta,
\end{equation}
Explicit forms of $r_n(x,\delta)$ exist and can be obtained under standard parametric, semiparametric, and nonparametric estimation frameworks. See, for example, \citet{vandervaart1998asymptotic} for regular parametric models, \citet{bickel1993efficient} for semiparametric estimation, and \citet{tsybakov2009introduction} for nonparametric estimation. We keep our analysis agnostic to model choice and only require that, for each fixed $\delta$, $r_n(x,\delta)\to0$ as $n\to\infty$.

Let $\Pbm=m^{-1}\sum_i\delta_{s_i}$ denote the empirical distribution constructed from i.i.d. source scenarios independent of calibration data, and let
\begin{equation}
    \label{eq:udt-hat}
    \widehat{\mathcal U}_{x}^{\rm DT}(\varepsilon)
    :=
    (\widehat T_x)_\#
    \left\{
    \Q:
    \W(\Q,\P_{b,x}^{(m)})\le \varepsilon
    \right\}
\end{equation}
denote the empirical ambiguity set used in DT-DRO.

\begin{theorem}[Radius for distributional coverage]
    \label{thm:radius}
Under Assumption~\ref{ass:diagnostic}, set
    \begin{equation}
    \label{eq:radius}
     \varepsilon=\D
     \left[r_n(x,\alpha/2)+\sqrt{\frac{\log(4/\alpha)}{2m}}\right],
    \end{equation}
then for $\widehat{\mathcal U}_x^{\rm DT}(\cdot)$ defined in \eqref{eq:udt-hat}, there is
    \begin{equation*}
        \Pr\left\{ \mathbb P_x^\star \in \widehat{\mathcal U}_x^{\rm DT}(\varepsilon) \right\} \geq 1- \alpha.
    \end{equation*}
\end{theorem}

\Cref{thm:radius} presents the analytical form of a valid radius choice, consisting of two additive components: The first term captures residual calibration error and vanishes as $\widehat G_x$ approaches the population diagnostic map. The second term is purely computational and arises from approximating $\widehat P_x$ with an empirical distribution based on $m$ samples. In practice, $r_n$ can be estimated and used in Algorithm~\ref{alg:dtdro} following a bootstrap-based procedure, developed in Appendix~\ref{app:radius}.

As the calibration sample and scenario count grow ($n,m\to\infty$), both diagnostic-estimation and scenario-approximation errors diminish, causing the valid radius to vanish asymptotically and the ambiguity set to contract. This contraction, in turn, leads to vanishing excess decision risk, as discussed next.

\subsection{Decision risk certificate}

For decision $z$ and set $\mathcal U$, define the robust upper risk
\[
    R(z;\mathcal U)
    \coloneqq
    \sup_{\Q\in \mathcal U}\int_{\cY} \ell(z,y)\,\Q(\dd y).
\]
Given the empirical DT-DRO ambiguity set, define the robust value and corresponding decision by
\begin{equation}
\label{eq:robust-value-estimator}
    \widehat z \in
        \argmin_{z\in\cZ} R\left(z;\widehat{\mathcal U}_{x}^{\rm DT}(\varepsilon) \right),
        \
        \widehat v\coloneqq R\left(\widehat z;\widehat{\mathcal U}_{x}^{\rm DT}(\varepsilon) \right)
\end{equation}
Note that the quantity $\widehat v$ is the finite-sample robust objective computed from the sample average approximated dual reformulation using the radius in \eqref{eq:radius}. Furthermore, let $\mathcal L^\star(z) = \int_\cY \ell(z, y) \Ps(\d y)$ and define the population-optimal value as
\begin{equation}
    \label{eq:pop-opt-val}
    v^\star \coloneqq \min_{z \in \mathcal{Z}} \mathcal{L}^\star(z).
\end{equation}

\begin{assumption}[Decision regularity]
    \label{ass:loss}
The feasible set $\cZ$ is nonempty and compact, $\ell$ is continuous in $(z,y)$, and there exists $L_y<\infty$ such that
    \begin{equation}
        \label{eq:loss-lipschitz-theory}
        \abs{\ell(z,y)-\ell(z,y')}
        \le L_y\abs{y-y'}
    \end{equation}
for all $z\in\cZ$ and $y,y'\in\cY$.
\end{assumption}

\begin{theorem}[Risk certificate and excess decision loss]
    \label{thm:value-certificate}
Let $\widehat z$, $\widehat v$ be defined in \eqref{eq:robust-value-estimator}, $v^\star$ be defined in \eqref{eq:pop-opt-val}. For $\varepsilon$ defined in \eqref{eq:radius}, write
    $$
     \omega_T(h)=\sup_{|s-s'|\le h}|T(s)-T(s')|,\quad B_{n,m}=\varepsilon+\overline\omega_{\Th}(\varepsilon),
    $$
where $\overline\omega_T$ is the least nondecreasing concave majorant of $\omega_T$, extended constantly beyond $\D$. Under Assumptions~\ref{ass:diagnostic} and \ref{ass:loss}, with probability at least $1-\alpha$,
    \begin{equation}
    \label{eq:regret}
     0\le
     \mathcal{L}^\star(\widehat z) - v^\star
     \le\widehat v-v^\star\le L_yB_{n,m}.
    \end{equation}
\end{theorem}

\Cref{thm:value-certificate} shows that the DT-DRO objective provides a valid high-probability certificate for the true risk of the selected decision. Specifically, $\widehat v$ upper-bounds the realized population risk $\mathcal{L}^\star(\widehat z)$, while the gap between this certificate and the population-optimal value $v^\star$ is controlled by $L_y B_{n,m}$. The term $B_{n,m}$ can be interpreted as the effective ambiguity width after accounting for both the statistical radius $\varepsilon$ and the distortion introduced by the fitted transport through $\overline\omega_{\widehat T}(\varepsilon)$. Thus, as the radius and transport-induced distortion shrink, both the conservatism of the robust objective and the excess risk of the decision vanish.

\begin{figure*}[t]
    \centering
    \includegraphics[width=1.0\linewidth]{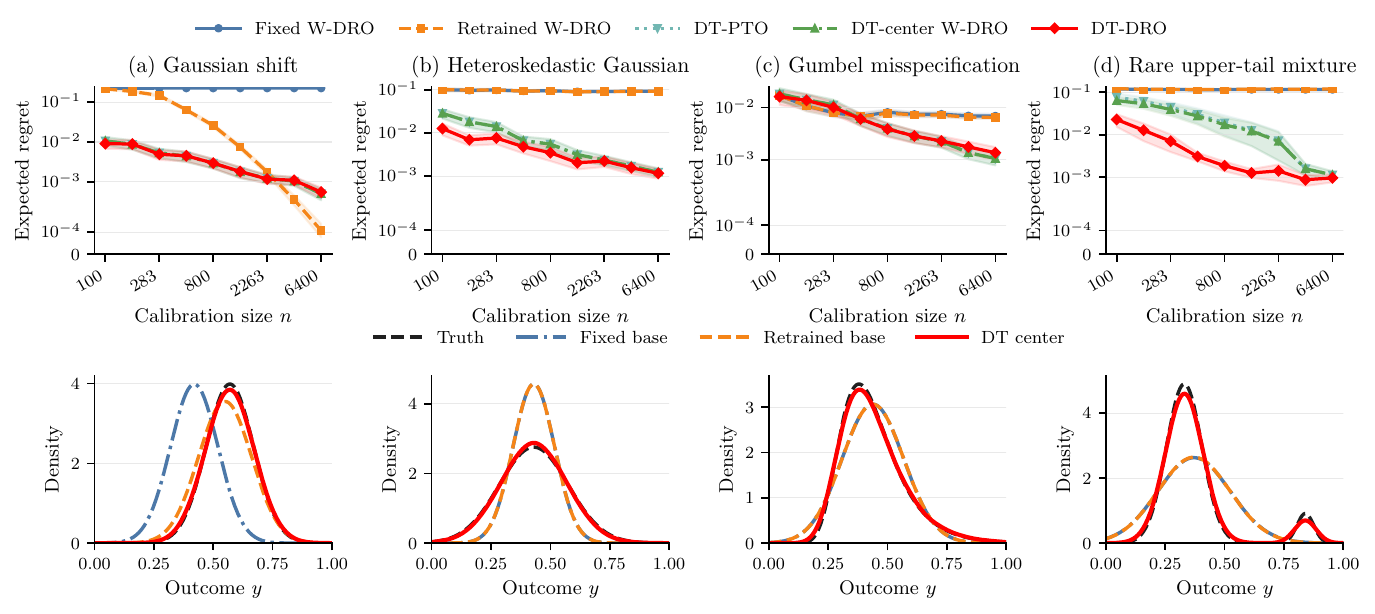}
    \caption{
Performance under four forms of predictive misspecification. \textit{Top row}: mean expected regret at the target context as the calibration-sample size increases, with shaded regions indicating $95\%$ confidence intervals over $100$ replications. \textit{Bottom row}: the true conditional density and mean fitted densities from different baselines. }
    \label{fig:synthetic}
\end{figure*}

\subsection{Statistical benefits of calibration}
\label{sec:comparisons}

The diagnostic wrapper provides a clear statistical benefit compared to model refitting schemes when the base model is misspecified. Model refitting refers to using the calibration data $\mathcal{D}$ to train the base model $\Pb$ from scratch \citep{kannan2024residuals,zhu2026decisiondependent}. It is effective when refitting can recover the true conditional distribution, but retains a persistent uncertainty floor otherwise. To formalize this, let $\widetilde\P_{n,x}$ denote the predictive distribution obtained after refitting on the available data, and suppose
$$
\W(\widetilde\P_{n,x},\P_x^\dagger)=o_p(1),
\quad
\Delta_{\rm refit}(x)
:=
\W(\P_x^\dagger,\Ps)>0.
$$
By the reverse triangle inequality,
\[
    \W(\widetilde\P_{n,x},\Ps)
    =
    \Delta_{\rm refit}(x)+o_p(1).
\]
Hence, any asymptotically valid Wasserstein ball centered at $\widetilde\P_{n,x}$ must have radius
\[
    \varepsilon_n^{\rm refit}
    \ge
    \Delta_{\rm refit}(x)+o_p(1).
\]
Moreover, under Assumption~\ref{ass:loss}, the corresponding Lipschitz decision-certificate width can be written as
\[
    \Gamma_n^{\rm refit}
    :=
    L_y\!\left[
        \varepsilon_n^{\rm refit}
        +
        \W(\widetilde\P_{n,x},\Ps)
    \right].
\]
For an asymptotically tight valid radius, the two quantities converge to $\Delta_{\rm refit}(x)$ and $2L_y\Delta_{\rm refit}(x)$, respectively. Thus, without altering the misspecified predictive model, increasing the sample size alone cannot eliminate the robustness floor induced by its systematic distributional error.

DT-DRO instead retains the predictive structure already encoded in the base model and learns only its residual distributional discrepancy through the diagnostic map $G_x^\star$. Because this correction acts as a wrapper rather than requiring the original model class to be redesigned, it can flexibly correct systematic errors that persist under repeated refitting (\Cref{thm:radius} and \Cref{thm:value-certificate}). Appendix~\ref{sec:geometry-ablation} further isolates the contribution of geometry calibration by holding the diagnostically corrected center fixed and comparing the transport-induced ground cost with the original Wasserstein ground cost.

\section{Numerical Experiments}
\label{sec:experiments}

We expand our analysis in \Cref{sec:comparisons} by numerically evaluate the performance of DT-DRO against four data-driven decision-making baselines: ($i$) \textit{Fixed W-DRO} applies standard Wasserstein DRO to the predictive distribution produced by the fixed base model; ($ii$) \textit{Retrained W-DRO} periodically refits the base model using all available historical observations instead of reserving samples for calibration before applying ordinary Wasserstein DRO; ($iii$) \textit{DT-PTO} makes a nominal predict-then-optimize decision using the diagnostically transported distribution without additional distributional robustness; and ($iv$) \textit{DT-center W-DRO} centers an ordinary Wasserstein ball at the diagnostically transported distribution, thereby isolating the benefit of center correction from that of the diagnostic-transport geometry used by DT-DRO. Additional details are provided in Appendix~\ref{app:exp}.

\subsection{Four synthetic data settings}

We consider four data-generation settings designed to capture distinct forms of predictive misspecification: ($i$) a Gaussian location shift between the base-training and deployment populations, ($ii$) a conditionally Gaussian distribution with a localized increase in variance around the target context, which cannot be captured by the homoskedastic base model, ($iii$) a heavy-tailed Gumbel outcome distribution, and ($iv$) a Gaussian mixture containing a rare upper-tail component. The base model in all four settings is fitted as a Gaussian GLM that either fails to capture the mode, variance, or tail behavior of the true underlying distribution from which $y\mid x$ is drawn, and the diagnostic transport is estimated from an independent calibration sample using a boundary-corrected local-neighbor KDE estimator of the conditional PIT-CDF. We evaluate decisions at a fixed target context over increasing calibration-sample sizes while using an independent validation sample to select the ambiguity radius to target 90\% empirical coverage.

\Cref{fig:synthetic} shows that DT-DRO generally benefits from increasing calibration data, with expected regret decreasing across all misspecification settings. Under a simple Gaussian shift, retraining eventually becomes competitive at large sample sizes, suggesting that refitting can be effective when misspecification is primarily parametric and the model family remains adequate. In contrast, under heteroskedastic, Gumbel, and rare upper-tail misspecification, retraining the same Gaussian model cannot fully correct the distributional error, whereas the diagnostic-transport methods consistently achieve lower regret and produce centers that more closely track the true distribution. The advantage is most pronounced for the rare upper-tail mixture, where DT-DRO substantially improves over DT-center W-DRO, indicating that adapting the ambiguity-set geometry provides benefits beyond merely correcting its center. Overall, the results suggest that diagnostic transport is especially valuable when misspecification is structural and cannot be resolved by conventional model retraining alone.

\subsection{Outage resilience planning}

\begin{figure}[t]
    \centering
    \includegraphics[width=1.0\linewidth]{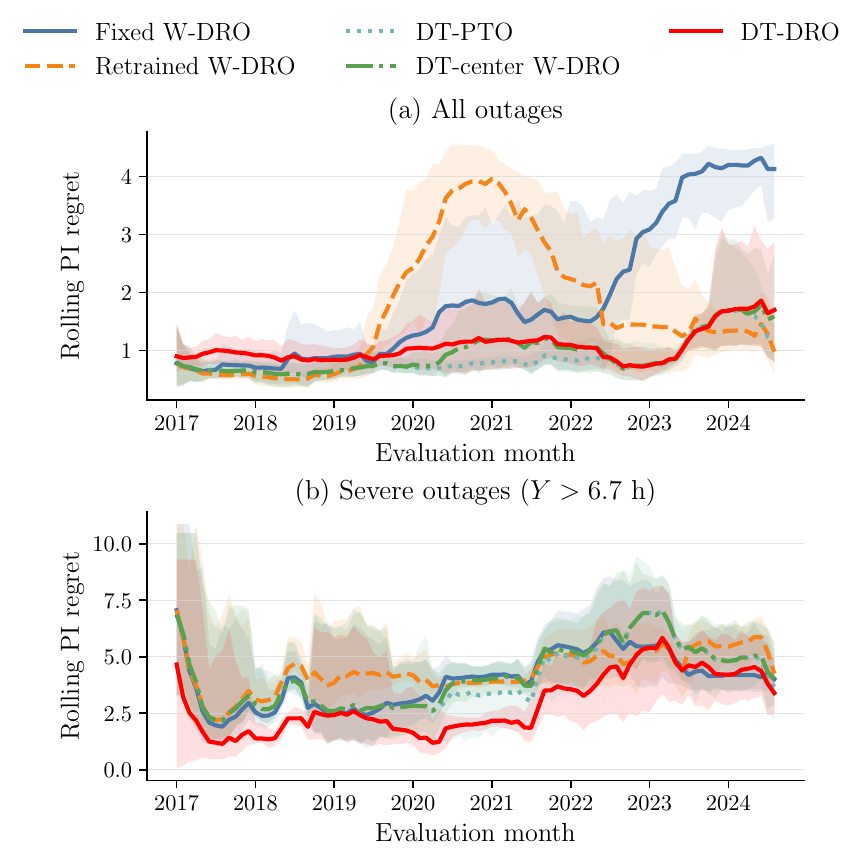}
    \caption{Rolling perfect-information (PI) regret in the sequential outage experiment for (a) all outages and (b) severe outages, defined as durations exceeding the weighted 95th percentile of the static training sample. Curves report the 12-month rolling weighted mean regret for each decision method, with shaded bands denoting $95\%$ confidence intervals from 300 substation-cluster bootstrap replications.}
    \label{fig:real}
\end{figure}

We further study a sequential procurement--shortage decision problem using real substation-level power-outage records from 2014--2024 provided by a proprietary electric utility in Indianapolis, Indiana. Outage duration, capped at 48 hours, serves as the response, with lagged weather, infrastructure, maintenance spending, seasonal effects, and substation indicators used as covariates. The fixed base distribution is a Gamma GLM with a log link fitted to the 2014--2016 static sample. Beginning in January 2017, decisions are evaluated monthly using only strictly prior observations: the retrained baseline refits the Gamma GLM on the expanding history, while the diagnostic transport is updated through a monotone gradient-boosted estimator of the conditional PIT-CDF. We compare against the same baselines as in the synthetic experiments and calibrate each method's ambiguity-radius multiplier using the preceding 12 evaluation months to target $90\%$ empirical coverage.

\Cref{fig:real} reports the 12-month rolling weighted perfect-information regret for both all outages and severe outages above the weighted 95th-percentile duration. For all outages, Fixed W-DRO becomes increasingly conservative over time, while periodic retraining reduces this deterioration but remains sensitive to temporal distribution shifts; in contrast, the diagnostic-transport methods maintain substantially lower and more stable regret over most of the evaluation horizon. The improvement persists when evaluation is restricted to severe outages above the historical $95\%$ duration threshold, indicating that the learned transport remains effective in the upper-tail where decision errors are most costly. DT-DRO generally performs at least as well as DT-PTO and DT-center W-DRO, suggesting that combining center correction with the transport-induced ambiguity geometry provides additional robustness beyond recalibration alone. Overall, the results indicate that diagnostic transport can adapt a fixed predictive model to evolving outage-duration distributions and improve robust decisions without requiring continual refitting of the base model.

\section{Conclusion}
\label{sec:conclusion}

We introduced diagnostic-transport DRO (DT-DRO), which uses calibration data to reshape an existing optimal transport ambiguity set without retraining the underlying predictive model. By estimating a diagnostic transport from the conditional probability integral transform cumulative distribution function, DT-DRO jointly corrects the ambiguity-set center and ground cost while preserving a tractable dual reformulation. We establish valid ambiguity radii and decision-risk certificates that shrink with diagnostic-estimation and scenario-approximation errors, and show that this can avoid the persistent conservatism caused by refitting a systematically misspecified model. Synthetic and power-outage experiments further demonstrate improved decision performance, particularly under structural misspecification.

A limitation of this work is that the learned transport $T_x$ enters the dual jointly with the loss, so our current exact finite-dimensional LP reformulation requires $T_x$ to be piecewise affine. Therefore, more general transport classes may require approximation to obtain a finite-dimensional optimization structure or resort to alternative numerical schemes, raising a tradeoff between diagnostic flexibility and computational tractability. A promising direction for future work is to extend DT-DRO to multivariate outcomes using copula-based constructions, allowing the framework to separately calibrate marginal predictive errors and cross-outcome dependence while preserving a tractable representation of the joint ambiguity set.


\bibliographystyle{apalike}
\bibliography{ref}

\clearpage
\onecolumn
\appendix
\begin{center}
{\Large\bfseries Appendices}
\end{center}

\section{Bootstrap estimation of the ambiguity radius}
\label{app:radius}

We first describe the practical radius estimate used to approximate the diagnostic-error term in \eqref{eq:radius}. Throughout, we hold the fitted base predictor fixed and use the calibration pairs $\mathcal D_{\rm cal}=\{(X_i,Y_i)\}_{i=1}^n$. Their PIT values are $U_i=F_b(Y_i\mid X_i)$, and the complete diagnostic fitting procedure returns an increasing estimate $\widehat G_x$. The purpose of the bootstrap is to assess how much this fitted curve changes when the calibration sample changes. This assessment is separate from the error introduced by generating finitely many source scenarios.

\subsection{Diagnostic variability and smoothing bias}
\label{app:radius-practice}

For each bootstrap replication $b=1,\ldots,B_{\rm boot}$, sample $n$ indices independently with replacement from $\{1,\ldots,n\}$ and refit the diagnostic using the selected pairs $(X_i,U_i)$. We resample entire calibration observations, not the expanded binary labels at different probability thresholds: labels constructed from one observation share the same PIT value. The base model is not retrained. Each bootstrap fit includes the same endpoint constraints, monotonicity treatment, and interpolation used by the original estimator. Smoothing parameters selected on an independent split can be held fixed; when parameter selection is treated as part of the estimator, that selection must be repeated within the bootstrap fit.

At the deployment context $x$, compute
\begin{equation}
\label{eq:bootstrap-deviation}
 q_b(x)=\sup_{u\in[0,1]}\left|\widehat G_x^{(b)}(u)-\widehat G_x(u)\right|,
 \qquad b=1,\ldots,B_{\rm boot}.
\end{equation}
Let $q_{(1)}(x)\le\cdots\le q_{(B_{\rm boot})}(x)$ be the ordered deviations. For diagnostic miscoverage allocation $\delta_{\rm cal}$, define the empirical bootstrap quantile by
\begin{equation}
\label{eq:bootstrap-quantile}
 \widehat q^{\rm boot}_{1-\delta_{\rm cal}}(x)
 :=q_{(\lceil B_{\rm boot}(1-\delta_{\rm cal})\rceil)}(x).
\end{equation}
For piecewise-affine CDF estimates, the supremum in \eqref{eq:bootstrap-deviation} is attained at an endpoint or a knot in the union of the two curves' knot sets. For other representations, evaluation on a finite probability grid approximates this supremum; any available interpolation bound should be included in the error allowance.

Bootstrap variation measures fluctuation around the fitted diagnostic and does not, by itself, account for its systematic smoothing bias. When the estimator has a bandwidth $h$, we use the sensitivity of the fit to halving that bandwidth as a practical bias allowance:
\begin{equation}
\label{eq:bootstrap-bias}
 \widehat b_x
 :=\left\|\widehat G_{x,h}-\widehat G_{x,h/2}\right\|_\infty+\eta_{\rm num},
\end{equation}
where $\eta_{\rm num}\ge0$ includes any separately controlled numerical approximation error not already represented by the fitted curves. The two fits use the same calibration observations.

\subsection{Adding the scenario-approximation term}

For $m$ independent source scenarios from $\Pb$, we allocate $\delta_{\rm cal}=\alpha/2$ to diagnostic estimation and the remaining $\alpha/2$ to the empirical source CDF, giving
\begin{equation}
 q_m=\sqrt{\frac{\log(4/\alpha)}{2m}},
 \qquad
 \widehat\varepsilon^{\rm proxy}_{n,m}(x)
 =\D\left[\widehat b_x+\widehat q^{\rm boot}_{1-\alpha/2}(x)+q_m\right].
\label{eq:radius-proxy}
\end{equation}
The three terms respectively account for smoothing bias, diagnostic sampling variability, and scenario approximation. The last term is the same analytical term as in \eqref{eq:radius}; only the diagnostic-error bound is replaced by a data-driven approximation.

When the source representation instead uses midpoint quantiles $s_i=F_b^{-1}((i-1/2)/m\mid x)$, its CDF error is at most $1/(2m)$ deterministically. We then use $q_m=1/(2m)$ and allocate all miscoverage to the diagnostic, replacing the bootstrap level $1-\alpha/2$ by $1-\alpha$. \Cref{app:containment} establishes both scenario bounds. Midpoint scenarios are deterministic and should not be described as independent random draws.

\begin{algorithm}[t]
\caption{Bootstrap proxy for the DT-DRO radius}
\label{alg:bootstrap-radius}
\begin{algorithmic}[1]
\Require Fitted base CDF $F_b$, calibration pairs $\mathcal D_{\rm cal}$, context $x$, scenario count $m$, target $\alpha$, bootstrap count $B_{\rm boot}$, and a specified diagnostic fitting procedure.
\State Compute $U_i=F_b(Y_i\mid X_i)$ and fit $\widehat G$ on $\{(X_i,U_i)\}_{i=1}^n$.
\State Set $(\delta_{\rm cal},q_m)=(\alpha/2,\sqrt{\log(4/\alpha)/(2m)})$ for independent source draws, or $(\alpha,1/(2m))$ for midpoint source quantiles.
\For{$b=1,\ldots,B_{\rm boot}$}
    \State Resample $n$ calibration pairs with replacement and refit the complete diagnostic procedure to obtain $\widehat G^{(b)}$.
    \State Evaluate $q_b(x)$ according to \eqref{eq:bootstrap-deviation}.
\EndFor
\State Compute $\widehat q^{\rm boot}_{1-\delta_{\rm cal}}(x)$ according to \eqref{eq:bootstrap-quantile}.
\State Specify $\widehat b_x$; for bandwidth-based fitting, use the proxy in \eqref{eq:bootstrap-bias}.
\State \Return $\widehat\varepsilon^{\rm proxy}_{n,m}(x)=\D[\widehat b_x+\widehat q^{\rm boot}_{1-\delta_{\rm cal}}(x)+q_m]$.
\end{algorithmic}
\end{algorithm}

\subsection{Theoretical guarantees of the proxy}

\Cref{thm:radius} remains valid with a data-dependent width $\widehat r_n(x,\delta)$ whenever that width satisfies
\begin{equation}
\label{eq:certified-random-width}
 \Pr\!\left\{\|\widehat G_x-G_x^\star\|_\infty\le\widehat r_n(x,\delta)\right\}\ge1-\delta.
\end{equation}
Indeed, the containment proof only needs this accuracy event. However, neither the bootstrap quantile nor the bandwidth-difference allowance automatically establishes \eqref{eq:certified-random-width} for an arbitrary learner. The bootstrap radius is therefore a practical proxy unless an estimator-specific bias and bootstrap-coverage argument is available. A finite number of bootstrap replications also introduces quantile-estimation error. We do not identify its empirical quantile with an exact finite-sample confidence bound.

A distinct, certified implementation is possible when an analytical diagnostic confidence width is available. \Cref{app:nonparametric} gives one local empirical-CDF construction with an explicit width when a valid smoothness bound is supplied. The experiments documented in \Cref{app:exp} use their own validation-based radius rules rather than \Cref{alg:bootstrap-radius}. Empirical coverage of a realized or group-average loss is useful for tuning those rules, but is not the same event as containment of the entire conditional distribution.

\section{Discussion of the regularity assumptions}
\label{app:ass}

\Cref{ass:diagnostic} combines support, invertibility, and sampling conditions. These conditions serve different roles, so relaxing one does not necessarily require changing every part of the framework. The set-level pushforward construction and its statistical calibration should, in particular, be distinguished.

\paragraph{Bounded support.}
For scalar distributions supported on $[a,b]$, the identity $\W(\P_1,\P_2)=\int_a^b|F_1(y)-F_2(y)|\,\dd y$ gives the bound $\W(\P_1,\P_2)\le\D\|F_1-F_2\|_\infty$. This is the step that turns the probability-scale diagnostic error into the radius in \eqref{eq:radius}. Bounded support is not needed to define a pushforward, but removing it requires control of the tails in both the outcome and source coordinates. A uniform CDF bound alone does not control their integrals over an unbounded domain. Truncation with a separately bounded tail remainder, or a direct transport-distance confidence bound, would replace the factor $\D$ argument. The squared-displacement objective in \eqref{eq:center-ot} would also require appropriate integrability.

\paragraph{Monotonicity and invertibility.}
The base CDF and the fitted diagnostic determine an increasing homeomorphism $\Th=F_b^{-1}\circ\Gh^{-1}\circ F_b$. Its inverse makes the induced ground cost in \eqref{eq:metric} well-defined and allows source and transported coordinates to be interchanged exactly. Strict monotonicity need not mean that a density is uniformly bounded away from zero. In fact, the containment proof compares CDFs at a common probability argument and does not require a quantitative bound on an inverse density. For a continuous nondecreasing fitted curve with endpoints zero and one, mixing it with the identity, $\widehat G_{x,\zeta}(u)=(1-\zeta)\widehat G_x(u)+\zeta u$ with $\zeta>0$, enforces strict increase and changes its uniform error by at most $\zeta$. A discontinuous empirical CDF must first be continuously interpolated; identity mixing alone does not remove jumps.

The population diagnostic $G_x^\star$ is not inverted in the finite-sample containment proof, so its strict monotonicity is stronger than that proof needs. It is used by the population quantile-map construction and by the uniform inverse-convergence argument underlying the asymptotic conclusion. Generalized inverses permit broader scalar calibration constructions, but flat regions or atoms can destroy the bijection required for the current induced-cost identity. Discrete outcomes may require a transport plan or a randomized map rather than a deterministic bijection. These extensions cannot be obtained merely by replacing ordinary inverses with generalized inverses in every theorem.

\paragraph{Sampling.}
The theory is conditional on the fitted base predictor. Independent target-population calibration pairs provide one transparent route to a diagnostic confidence bound, and independence from the base-training sample makes the conditional analysis straightforward. Once a valid event of the form \eqref{eq:maprate} is available, the containment argument does not otherwise depend on how the diagnostic was fitted. Dependent or clustered calibration data can therefore be accommodated only after establishing a suitable diagnostic error bound for that sampling structure. Resampling time blocks or clusters may be useful computationally, but does not by itself establish the required confidence statement. The rolling outage experiment uses past observations and weighted clusters; its empirical evaluation is consequently separate from the independent-calibration guarantee.

\paragraph{Convergence and computation.}
Finite-sample containment uses an accuracy bound at the chosen confidence level; asymptotic contraction additionally requires that bound and the scenario error to vanish. Transferring a linear-in-radius rate to outcome-space risk further requires control of the fitted transport's Lipschitz constants, whereas consistency can follow from convergence to a continuous population transport. Finally, piecewise-affine transport is a sufficient condition for the exact LP reduction in \Cref{cor:lp-reduction}, not a necessary condition for all possible finite reformulations. More general maps leave a separation or approximation problem inside the dual, as discussed in \Cref{app:lp}.

\section{Detailed proofs}
\label{app:proofs}

We fix the deployment context $x$ and condition on the fitted base predictor throughout. Unless stated otherwise, the scalar distributions are supported on $\cY=[a,b]$. We write $F_b$ for $F_b(\cdot\mid x)$ when the context is unambiguous, and distinguish the predictive law $\Pb$, its source-scenario approximation $\Pbm$, and the corrected empirical center $\Phm=(\Th)_\#\Pbm$. To distinguish evaluation under a fixed distribution from the robust upper risk $R(z;\mathcal U)$ defined in the main text, we write $\mathcal L(z;\Q)\coloneqq\int_{\cY}\ell(z,y)\,\Q(\dd y)$; hence $\mathcal L^\star(z)=\mathcal L(z;\Ps)$. The fitting and scenario errors enter the statistical proofs through separate events.

\subsection{Proof of \texorpdfstring{\Cref{prop:ball}}{Lemma 1} and the dual reformulation}
\label{app:ball}

Let $T=T_x$ be the measurable bijection in \Cref{prop:ball}. For any coupling $\pi$ of source laws $\Q_1$ and $\Q_2$, the pushforward $(T,T)_\#\pi$ couples $T_\#\Q_1$ and $T_\#\Q_2$, and
\begin{equation}
 \int c_T(y,y')\,(T,T)_\#\pi(\dd y,\dd y')
 =\int c_T(T(s),T(s'))\,\pi(\dd s,\dd s')
 =\int c(s,s')\,\pi(\dd s,\dd s').
\end{equation}
Conversely, every coupling of the two transported laws pulls back under $(T^{-1},T^{-1})$ to a source coupling with the same cost. The correspondence is bijective, so taking infima yields
\begin{equation}
\label{eq:appisometry}
 \mathsf D_{c_T}(T_\#\Q_1,T_\#\Q_2)=\mathsf D_c(\Q_1,\Q_2).
\end{equation}
Each transported candidate $\Q$ has the unique preimage $(T^{-1})_\#\Q$. Applying \eqref{eq:appisometry} with $\Q_2=\Pb$ proves \eqref{eq:ball}. This geometric identity uses bijectivity and measurability; it does not require estimating $T$ by a particular method. Pushforward also preserves mixtures, since $T_\#(t\Q_1+(1-t)\Q_2)=tT_\#\Q_1+(1-t)T_\#\Q_2$ for $t\in[0,1]$.

For the duality step we additionally use the usual regularity conditions for OT strong duality \citep{gao2023distributionally,blanchet2022optimal}. A sufficient version for the compact scalar setting is that $c$ is continuous and nonnegative with $c(s,s)=0$, and that $T$ and $\ell(z,\cdot)$ are continuous for each fixed $z$. These conditions make the following coupling problem compact and its objectives continuous.

Fix $z$ and put $f_z(s)=\ell(z,T(s))$. By pushforward integration, the worst-case risk is
\begin{equation}
\label{eq:app-source-primal}
 R\left(z;\cU_x^{\rm DT}(\varepsilon)\right)
 =\sup_{\Q\in\cU_x^{\rm DT}(\varepsilon)}\mathcal L(z;\Q)
 =\sup_{\mathsf D_c(\Q_0,\Pb)\le\varepsilon}\int f_z(s)\,\Q_0(\dd s)
 =\sup_{\substack{\pi_2=\Pb\\ \int c(s,S)\,\dd\pi\le\varepsilon}}\int f_z(s)\,\pi(\dd s,\dd S).
\end{equation}
Here $\pi_2$ denotes the second marginal. The Lagrange multiplier $\lambda\ge0$ for the transportation budget gives
\begin{equation}
\label{eq:app-source-dual}
 \inf_{\lambda\ge0}\left\{\lambda\varepsilon+\int\sup_{s\in\cY}\bigl[f_z(s)-\lambda c(s,S)\bigr]\,\Pb(\dd S)\right\}.
\end{equation}
For $\varepsilon>0$, the diagonal coupling has zero transportation cost, and the compact coupling problem satisfies the budget qualification for strong duality. For fixed $\lambda$, disintegration over the second marginal and measurable maximization on the compact outcome space give the integral of the pointwise supremum in \eqref{eq:app-source-dual}. At $\varepsilon=0$, use positive radii decreasing to zero: compactness of feasible couplings and continuity imply convergence of the primal values, and weak duality together with this limit gives the same infimum identity. A finite minimizing multiplier need not exist at radius zero.

Changing variables $u=T(S)$ and $y=T(s)$ in \eqref{eq:app-source-dual} gives the transported-coordinate expression in \eqref{eq:dual}. Taking the infimum over $z$ establishes the stated decision formulation; no interchange of the decision infimum with an adversarial supremum is needed. For computation, source draws $\nu_i\sim\Pb$ and transported draws $u_i=T(\nu_i)$ give identical empirical objectives:
\begin{equation}
\label{eq:app-coordinate-saa}
 \frac1m\sum_{i=1}^m\sup_y\bigl[\ell(z,y)-\lambda c_T(y,u_i)\bigr]
 =\frac1m\sum_{i=1}^m\sup_s\bigl[\ell(z,T(s))-\lambda c(s,\nu_i)\bigr].
\end{equation}
Thus the scenario distribution and the coordinates in the loss and cost must be changed together. Replacing the predictive center by its empirical scenario law is the sample-average approximation; dualizing the resulting empirical-center DRO problem is exact.

\subsection{Proof of \texorpdfstring{\Cref{cor:lp-reduction}}{Theorem 1}}
\label{app:lp}
\label{app:computation}

Work in the source coordinates of \eqref{eq:app-coordinate-saa}. For fixed $z$, $\lambda\ge0$, and scenario index $i$, define
\begin{equation}
 \phi_i(s)=\ell(z,T_x(s))-\lambda|s-\nu_i|,
 \qquad \mathcal K_i=\{\tau_0,\ldots,\tau_K,\nu_i\}.
\end{equation}
Sort the distinct points in $\mathcal K_i$. Between two consecutive points, $T_x(s)$ and $|s-\nu_i|$ are affine in $s$. Therefore, on that interval,
\begin{equation}
 \phi_i(s)=\max_{r\in[R]}\left\{\alpha_r^\top z+\beta_rT_x(s)+\gamma_r-\lambda|s-\nu_i|\right\}
\end{equation}
is a maximum of affine functions and hence convex. Its maximum on the closed interval is attained at an endpoint. Taking the union of all endpoints proves
\begin{equation}
\label{eq:app-finite-max}
 \sup_{s\in[a,b]}\phi_i(s)=\max_{v\in\mathcal K_i}\phi_i(v).
\end{equation}
It is unnecessary to add the loss's decision-dependent kinks to $\mathcal K_i$: convexity already places the maximum at the interval endpoints.

Introduce one epigraph variable $h_i$ for each maximum. The inequalities $h_i\ge\phi_i(v)$ are equivalent to
\begin{equation}
 h_i\ge\alpha_r^\top z+\beta_rT_x(v)+\gamma_r-\lambda|v-\nu_i|,
 \qquad v\in\mathcal K_i,\quad r\in[R],\quad i\in[m].
\end{equation}
For a fitted transport and fixed scenarios, $T_x(v)$ and $|v-\nu_i|$ are constants. The objective and all displayed constraints are linear in $(z,\lambda,h)$, and $\cZ$ is polyhedral, proving \eqref{eq:finite-lp}. There are $\dim(z)+m+1$ scalar variables and at most $mR(K+2)$ loss inequalities, in addition to the description of $\cZ$ and $\lambda\ge0$. The reduction is exact for the scenario approximation and the fitted piecewise-affine map, not a replacement of a general population integral by a finite sum without approximation.

\paragraph{Approximating a more general transport.}
We also control the effect of replacing $T$ by a numerical approximation $\widetilde T$. If $\|T-\widetilde T\|_\infty\le\eta$ and the loss is uniformly $L_y$-Lipschitz in its outcome, then, for every source law $\Q_0$ and decision $z$,
\begin{equation}
 \left|\E_{\Q_0}\ell(z,T(S))-\E_{\Q_0}\ell(z,\widetilde T(S))\right|\le L_y\eta.
\end{equation}
Taking suprema over the same source ambiguity set preserves this bound, as does taking infima over $z$. The optimal values differ by at most $L_y\eta$, and a decision optimized using $\widetilde T$ is at most $2L_y\eta$ suboptimal for the robust objective using $T$. This is an optimization-approximation statement. Distributional containment for a modified transport still needs its own diagnostic-error allowance. Similarly, restricting a general inner supremum to a finite grid can underestimate worst-case risk unless a separate approximation bound is supplied.

\subsection{Proof of \texorpdfstring{\Cref{prop:minimum-displacement}}{Lemma 2}}
\label{app:diagnostic-proof}

The conditional PIT identity and the associated recalibration construction follow the diagnostic-transport literature \citep{zhao2021diagnostics,dey2025ladar,cucuzzella2026trustworthy}. We give the details in our notation and connect the construction explicitly to the squared-displacement objective in \eqref{eq:center-ot}.

Strict monotonicity of $F_b$ gives $\{Y\le y\}=\{F_b(Y)\le F_b(y)\}$ conditional on $X=x$. Consequently,
\begin{equation}
\label{eq:app-population-identity}
 F_x^\star(y)=G_x^\star(F_b(y)),
 \qquad
 (F_x^\star)^{-1}=F_b^{-1}\circ(G_x^\star)^{-1}.
\end{equation}
The candidate map in \eqref{eq:tx-oracle} is therefore $T_x^\star=(F_x^\star)^{-1}\circ F_b$. For $B\sim\Pb$, continuity of the base CDF implies $F_b(B)\sim\mathrm{Unif}(0,1)$, and hence
\begin{equation}
 \Pr\{T_x^\star(B)\le y\}=\Pr\{F_b(B)\le F_x^\star(y)\}=F_x^\star(y).
\end{equation}
Thus $(T_x^\star)_\#\Pb=\Ps$, so the candidate is feasible.

To prove minimum displacement, suppress $x$ and write $T=T_x^\star$, $\mu=\Pb$, and $\nu=\Ps$. Let $S$ be any other feasible measurable map. Since $S(B)$ and $T(B)$ both have law $\nu$, their second moments agree. The two costs therefore differ by
\begin{equation}
\label{eq:ot-proof-cost-gap}
 J(S)-J(T)=2\left(\E[BT(B)]-\E[BS(B)]\right),
 \qquad J(S):=\E|S(B)-B|^2.
\end{equation}
All terms are finite on bounded support. Define $K(v)=\int_a^vT^{-1}(u)\,\dd u$. For $y,v\in[a,b]$,
\begin{equation}
 K(v)-K(T(y))-y(v-T(y))
 =\int_{T(y)}^v\bigl[T^{-1}(u)-y\bigr]\,\dd u\ge0,
\end{equation}
with equality exactly when $v=T(y)$, because $T^{-1}$ is strictly increasing. Substitute $y=B$ and $v=S(B)$ and take expectations. The $K$ terms cancel because the output laws agree, leaving $\E[BT(B)]-\E[BS(B)]\ge0$. Together with \eqref{eq:ot-proof-cost-gap}, this proves optimality. Equality of costs forces $S(B)=T(B)$ almost surely, so the optimal map is unique up to a base-law null set. Its cost is
\begin{equation}
\label{eq:minimum-displacement-value}
 J(T_x^\star)=\int_0^1\left|(F_x^\star)^{-1}(u)-F_b^{-1}(u\mid x)\right|^2\,\dd u.
\end{equation}
This minimizes displacement of outcome values subject to matching the output law; it does not minimize a Euclidean distance between curves in a P--P plot.

For later use, the fitted map $\Th=F_b^{-1}\circ\Gh^{-1}\circ F_b$ has inverse $\Th^{-1}=F_b^{-1}\circ\Gh\circ F_b$, and the CDF of its corrected population center is
\begin{equation}
\label{eq:app-corrected-cdf}
 \widehat F_x(y)=F_b(\Th^{-1}(y))=\widehat G_x(F_b(y)).
\end{equation}
This identity is exact for an admissible fitted diagnostic, whether or not it equals $G_x^\star$. In the population, $G_x^\star(u)=u$ is equivalent to $F_x^\star=F_b$ and then $T_x^\star$ is the identity map.

\subsection{Proof of \texorpdfstring{\Cref{thm:radius}}{Theorem 2}}
\label{app:containment}

Let $T=\Th$, let $F_{b,m}$ be the empirical CDF of $s_1,\ldots,s_m\sim\Pb$, and set $r=r_n(x,\alpha/2)$ and $q_m=\sqrt{\log(4/\alpha)/(2m)}$. The diagnostic accuracy bound \eqref{eq:maprate} and the Dvoretzky--Kiefer--Wolfowitz inequality with the sharp constant \citep{massart1990} imply that the event
\begin{equation}
\label{eq:app-coverage-event}
 \mathcal E=\left\{\|\Gh-\Gs\|_\infty\le r\right\}\cap\left\{\|F_{b,m}-F_b\|_\infty\le q_m\right\}
\end{equation}
has probability at least $1-\alpha$. The source scenarios are generated independently of calibration. A union bound combines the two error events; independence of those events is not otherwise needed for this step.

We establish containment in both the transported and the ordinary geometry at the corrected empirical center. Write $\widehat F=\Gh\circ F_b$ for the CDF of $\Ph=T_\#\Pb$ and $\widehat F_m=F_{b,m}\circ T^{-1}$ for that of $\Phm=T_\#\Pbm$. On $\mathcal E$,
\begin{equation}
 \|F_x^\star-\widehat F\|_\infty\le r,
 \qquad
 \|\widehat F-\widehat F_m\|_\infty\le q_m.
\end{equation}
The first inequality follows from \eqref{eq:app-population-identity} and \eqref{eq:app-corrected-cdf}; the second follows by composing the source CDFs with the same inverse map. Integrating over the bounded support gives
\begin{equation}
\label{eq:app-center-error}
 \W(\Ps,\Phm)=\int_a^b|F_x^\star(y)-\widehat F_m(y)|\,\dd y\le\D(r+q_m)=\varepsilon.
\end{equation}

For the transported geometry, let $H$ be the CDF of $(T^{-1})_\#\Ps$. Since $T$ is increasing,
\begin{equation}
 H(s)=F_x^\star(T(s))=G_x^\star\!\left(\widehat G_x^{-1}(F_b(s))\right).
\end{equation}
At this same probability argument, $F_b(s)=\widehat G_x(\widehat G_x^{-1}(F_b(s)))$, so $\|H-F_b\|_\infty\le r$ on $\mathcal E$. Thus
\begin{equation}
\label{eq:app-pullback-error}
 \mathsf D_{c_T}(\Ps,\Phm)
 =\W((T^{-1})_\#\Ps,\Pbm)
 =\int_a^b|H(s)-F_{b,m}(s)|\,\dd s
 \le\D(r+q_m)=\varepsilon.
\end{equation}
The first equality is \eqref{eq:appisometry}. Since $(T^{-1})_\#\Ps$ lies in the source ball in \eqref{eq:udt-hat}, its pushforward is an element of $\widehat{\mathcal U}_x^{\rm DT}(\varepsilon)$. This proves \Cref{thm:radius}. Equations \eqref{eq:app-center-error} and \eqref{eq:app-pullback-error} hold on the same event, which will also allow a common coverage comparison with center-only calibration.

\paragraph{Midpoint source quantiles.}
For $s_i=F_b^{-1}((i-1/2)/m\mid x)$, the empirical CDF equals $i/m$ between $s_i$ and $s_{i+1}$, while the continuous base CDF ranges between $(i-1/2)/m$ and $(i+1/2)/m$. The endpoint intervals satisfy the same error bound, and hence $\|F_{b,m}-F_b\|_\infty\le1/(2m)$ deterministically. Repeating the proof with diagnostic failure probability $\alpha$ gives
\begin{equation}
\label{eq:app-midpoint-radius}
 \varepsilon_{n,m}^{\rm mid}(x;\alpha)=\D\left[r_n(x,\alpha)+\frac1{2m}\right].
\end{equation}
Without scenario approximation, both distances to $\Ph$ are bounded by $\D\|\Gh-\Gs\|_\infty$. These statements do not require an inverse-CDF error bound: inversion only selects the argument at which $G_x^\star$ and $\widehat G_x$ are compared.

\subsubsection{One constructive diagnostic confidence bound}
\label{app:nonparametric}

The coverage theorem is conditional on an appropriate diagnostic accuracy bound. We retain one explicit construction to show how such a bound can be obtained without tying the framework to this particular estimator. Let $X\in\R^d$ and fix $x$. Suppose that, in a neighborhood of $x$, the conditional PIT CDFs obey
\begin{equation}
 \sup_{u\in[0,1]}|G_{x'}^\star(u)-G_x^\star(u)|\le L\|x'-x\|^\beta,
 \qquad 0<\beta\le1,
 \qquad
 p_h:=\Pr\{\|X-x\|\le h\}\ge c_0h^d
\end{equation}
for sufficiently small $h$, with $c_0>0$. These are additional conditions for this example, not requirements on every diagnostic learner used in DT-DRO.

For a deterministic bandwidth $h$, let $I_h=\{i:\|X_i-x\|\le h\}$, $N_h=|I_h|$, and, when $N_h>0$, define $G^{\rm emp}_{x,h}(u)=N_h^{-1}\sum_{i\in I_h}\1\{U_i\le u\}$. Construct a continuous strictly increasing estimate $\widehat G_{x,h}$ with $\|\widehat G_{x,h}-G^{\rm emp}_{x,h}\|_\infty\le\eta_h$. If a valid upper bound $\overline L\ge L$ is supplied, then a random confidence width is
\begin{equation}
\label{eq:localband}
 \widehat r_h(x,\delta)=
 \begin{cases}
 \displaystyle\min\left\{1,\overline Lh^\beta+\sqrt{\frac{\log(2/\delta)}{2N_h}}+\eta_h\right\},&N_h>0,\\[2mm]
 1,&N_h=0.
 \end{cases}
\end{equation}
It satisfies \eqref{eq:certified-random-width} for this estimator. To see this, condition on $N_h=k>0$ and on membership in the fixed neighborhood. The selected PIT values have the mixture CDF $G_h(u)=\E[G_X^\star(u)\mid\|X-x\|\le h]$, whose uniform bias relative to $G_x^\star$ is at most $Lh^\beta$. The DKW inequality bounds $\|G^{\rm emp}_{x,h}-G_h\|_\infty$ by $\sqrt{\log(2/\delta)/(2k)}$ with conditional probability at least $1-\delta$. The triangle inequality adds the bias and interpolation terms. Averaging over $k$ proves the bound, and width one is always valid on an empty neighborhood.

For continuous PIT laws, the selected order statistics $U_{(1)}<\cdots<U_{(k)}$ are distinct almost surely. Interpolating through $(0,0)$, $(U_{(j)},(j-1/2)/k)$ for $j=1,\ldots,k$, and $(1,1)$ produces an increasing homeomorphism with $\eta_h\le1/(2k)$. A further identity mixture of weight $\zeta_n$ changes the uniform error by at most $\zeta_n$.

For completeness, the binomial count bound $\Pr\{N_h<np_h/2\}\le\exp(-np_h/8)$ gives, when $np_h\ge8\log(2/\delta)$, the deterministic-order estimate
\begin{equation}
 \|\widehat G_{x,h}-G_x^\star\|_\infty
 \le Lh^\beta+\sqrt{\frac{\log(4/\delta)}{nc_0h^d}}+\eta_h
\end{equation}
with probability at least $1-\delta$. Balancing the first two terms gives $h\asymp(\log(cn/\delta)/n)^{1/(2\beta+d)}$ and rate $(\log(cn/\delta)/n)^{\beta/(2\beta+d)}$ when the interpolation error is no larger. A bandwidth selected from the same responses requires a simultaneous bound, an independent selection split, or another estimator-specific inference argument. This example does not establish that the different KDE and boosting procedures in the experiments have the same finite-sample confidence guarantee.

\subsection{Proof of \texorpdfstring{\Cref{thm:value-certificate}}{Theorem 3}}
\label{app:value-certificate}
\label{app:modulus}

The key step is to translate a source-space transportation budget into an outcome-space error. For a continuous map $T:[a,b]\to[a,b]$, recall $\omega_T(h)=\sup_{|s-s'|\le h}|T(s)-T(s')|$. Its least nondecreasing concave majorant has the representation
\begin{equation}
\label{eq:app-modulus-representation}
 \overline\omega_T(t)=\sup\left\{\E\omega_T(D):0\le D\le\D,\ \E D\le t\right\}.
\end{equation}
The right-hand side is nondecreasing, is concave by mixing feasible laws of $D$, and majorizes $\omega_T$ by taking $D=t$ for $t\le\D$. Conversely, every nondecreasing concave majorant bounds the right-hand side by Jensen's inequality. This proves the representation and gives the constant extension for $t>\D$.

Let $\pi$ be an optimal $\W$ coupling of $\Q_1$ and $\Q_2$. Transporting that coupling and applying Jensen's inequality gives
\begin{equation}
\label{eq:modulusbound}
 \W(T_\#\Q_1,T_\#\Q_2)
 \le\int\omega_T(|s-s'|)\,\dd\pi
 \le\overline\omega_T\!\left(\int|s-s'|\,\dd\pi\right)
 =\overline\omega_T(\W(\Q_1,\Q_2)).
\end{equation}
Moreover, for any $h>0$ and any $D$ feasible in \eqref{eq:app-modulus-representation},
\begin{equation}
 \E\omega_T(D)\le\omega_T(h)+\D\Pr\{D>h\}\le\omega_T(h)+\D t/h.
\end{equation}
First send $t$ to zero and then $h$ to zero. Uniform continuity of $T$ implies $\overline\omega_T(t)\to0$. If $T$ is $L_T$-Lipschitz, \eqref{eq:app-modulus-representation} also gives $\overline\omega_T(t)\le L_Tt$.

On the event $\mathcal E$ from \Cref{app:containment}, every $\Q\in\widehat{\mathcal U}_x^{\rm DT}(\varepsilon)$ can be written as $\Q=T_\#\Q_0$ with $T=\Th$ and $\W(\Q_0,\Pbm)\le\varepsilon$. Combining \eqref{eq:app-center-error} and \eqref{eq:modulusbound} yields
\begin{equation}
\label{eq:excessbound}
 \W(\Q,\Ps)\le\W(T_\#\Q_0,T_\#\Pbm)+\W(\Phm,\Ps)
 \le\overline\omega_T(\varepsilon)+\varepsilon=B_{n,m}.
\end{equation}
Thus $B_{n,m}$ bounds the farthest ordinary Wasserstein departure from the truth among all distributions retained by the transported set.

The loss regularity in \Cref{ass:loss} gives $|\mathcal L(z;\Q)-\mathcal L^\star(z)|\le L_y\W(\Q,\Ps)$ uniformly over $z$. Containment of $\Ps$ and \eqref{eq:excessbound} therefore imply
\begin{equation}
 \mathcal L^\star(z)\le R\left(z;\widehat{\mathcal U}_x^{\rm DT}(\varepsilon)\right)
 \le \mathcal L^\star(z)+L_yB_{n,m},\qquad z\in\cZ.
\end{equation}
Compactness of $\cZ$ and continuity of the loss give a population optimizer $z^\star$. They also ensure continuity of the robust objective: on the compact product $\cZ\times\cY$, the loss is uniformly continuous, and taking suprema of its expectations preserves this uniform bound. Hence a robust optimizer exists. Evaluating the preceding inequalities at $\widehat z$ and $z^\star$ gives
\begin{equation}
 v^\star\le \mathcal L^\star(\widehat z)\le\widehat v
 =R\left(\widehat z;\widehat{\mathcal U}_x^{\rm DT}(\varepsilon)\right)
 \le R\left(z^\star;\widehat{\mathcal U}_x^{\rm DT}(\varepsilon)\right)
 \le v^\star+L_yB_{n,m},
\end{equation}
which is \eqref{eq:regret}. The event has probability at least $1-\alpha$.

\subsubsection{Asymptotic interpretation and rates}
\label{app:consistency}

To justify contraction as sample sizes grow, take a sequence indexed by $k$ with $n_k,m_k\to\infty$, $\alpha_k\to0$, and $\varepsilon_k\to0$, where $\varepsilon_k$ is \eqref{eq:radius} at $(n_k,m_k,\alpha_k)$. In particular, $r_k=r_{n_k}(x,\alpha_k/2)\to0$. On the corresponding accuracy events, monotonicity implies
\begin{equation}
 (G_x^\star)^{-1}((u-r_k)\vee0)
 \le\widehat G_{x,k}^{-1}(u)
 \le(G_x^\star)^{-1}((u+r_k)\wedge1),\qquad u\in[0,1].
\end{equation}
Uniform continuity of $(G_x^\star)^{-1}$ and $F_b^{-1}$ therefore gives $\eta_k:=\|\widehat T_{x,k}-T_x^\star\|_\infty\to0$. Since $\omega_{\widehat T_{x,k}}(h)\le2\eta_k+\omega_{T_x^\star}(h)$, \eqref{eq:app-modulus-representation} gives
\begin{equation}
 \overline\omega_{\widehat T_{x,k}}(\varepsilon_k)
 \le2\eta_k+\overline\omega_{T_x^\star}(\varepsilon_k)\longrightarrow0.
\end{equation}
Consequently $B_{n_k,m_k}\to0$ on events whose probabilities tend to one. The robust value and the achieved population risk converge in probability to $v^\star$. Compactness and continuity imply that the decisions approach the population optimizer set: outside any fixed neighborhood of that set, population risk exceeds $v^\star$ by a strictly positive amount. A unique optimizer yields decision convergence. Summable $\alpha_k$ give the corresponding almost-sure conclusions by the first Borel--Cantelli lemma.

When the fitted transport Lipschitz constants are uniformly bounded by $L_T$, the simpler inequality $B_{n,m}\le(1+L_T)\varepsilon$ transfers the diagnostic radius rate directly to the risk bound. For example, the estimator-specific conditions in \Cref{app:nonparametric} give $B_{n,m}=\widetilde O(n^{-\beta/(2\beta+d)}+m^{-1/2})$ for fixed confidence, or confidence sequences whose logarithmic terms have the same order. Without slope control, the modulus still yields consistency under the preceding conditions, but not necessarily the same linear-in-radius rate.

\subsection{Extended comparison in \texorpdfstring{\Cref{sec:comparisons}}{Section 4.3}: center and geometry}
\label{sec:geometry-ablation}
\label{app:geometry}
\label{app:comparison}

We separate two mechanisms. Correcting a misspecified center can remove a persistent distributional error floor. Changing the ground cost can additionally reduce the robust premium after the corrected center has been fixed. The second comparison isolates geometry rather than attributing every benefit of recalibration to it.

\subsubsection{Persistent error under refitting}

Let $\widetilde\P_{n,x}$ be the refitted law in \Cref{sec:comparisons}, let $d_n=\W(\widetilde\P_{n,x},\Ps)$, and assume $\W(\widetilde\P_{n,x},\P_x^\dagger)=o_p(1)$ with $\Delta_{\rm refit}(x)=\W(\P_x^\dagger,\Ps)>0$. The reverse triangle inequality gives $|d_n-\Delta_{\rm refit}(x)|\le\W(\widetilde\P_{n,x},\P_x^\dagger)=o_p(1)$. Define the ordinary refitted ball $\mathcal B_n^{\rm refit}=\{\Q:\W(\Q,\widetilde\P_{n,x})\le\varepsilon_n^{\rm refit}\}$. On every event where this ball contains $\Ps$,
\begin{equation}
\label{eq:app-refit-floor}
 \varepsilon_n^{\rm refit}\ge d_n,
 \qquad
 \sup_{\Q\in\mathcal B_n^{\rm refit}}\W(\Q,\Ps)\ge d_n,
 \qquad
 \Gamma_n^{\rm refit}:=L_y(\varepsilon_n^{\rm refit}+d_n)\ge2L_yd_n.
\end{equation}
The middle inequality holds even without containment, because every such ball contains its own center. The triangle inequality and loss Lipschitz continuity show that $\Gamma_n^{\rm refit}$ is a valid upper bound on the robust-value gap on the containment event. If $\varepsilon_n^{\rm refit}-d_n\to0$ in probability, then the radius and this particular upper-bound width converge to $\Delta_{\rm refit}(x)$ and $2L_y\Delta_{\rm refit}(x)$, respectively.

The probability qualification matters: a $1-\alpha$ coverage guarantee gives the radius lower bound on a $1-\alpha$ event, not on all samples. When coverage tends to one, the lower-bound statement can be read asymptotically in probability. Also, a nonvanishing upper-bound width does not establish a positive lower bound on the actual regret or robust-value gap for every fixed loss. The distributional error floor is unavoidable for the stated limiting center; its effect on a particular decision remains loss-dependent.

In contrast, \Cref{app:consistency} gives vanishing excess ambiguity for the diagnostically transported set under consistent diagnostic estimation and vanishing scenario error. An ordinary Wasserstein ball at the same corrected empirical center also has excess ambiguity at most $2\varepsilon$ on $\mathcal E$. Thus removal of the refitting floor is a benefit of center correction; the additional benefit of the induced geometry is addressed next.

\subsubsection{How the ground cost changes distributional discrepancies}
\label{app:metric}

Fix an increasing homeomorphism $T$ with inverse $S$. For any outcome laws $\Q_1,\Q_2$, the induced scalar OT cost satisfies
\begin{equation}
\label{eq:app-cdf-geometry}
 \mathsf D_{c_T}(\Q_1,\Q_2)
 =\W(S_\#\Q_1,S_\#\Q_2)
 =\int_a^b|F_{\Q_1}(T(s))-F_{\Q_2}(T(s))|\,\dd s
 =\int_a^b|F_{\Q_1}(y)-F_{\Q_2}(y)|\,\dd S(y).
\end{equation}
The first equality is \eqref{eq:appisometry}; the second follows from the CDF of a pulled-back law; the last is a change of variables with Lebesgue--Stieltjes measure. The center-only construction integrates the same CDF discrepancy against $\dd y$, whereas DT-DRO uses $\dd S(y)$. This is the precise difference between moving the center and changing the geometry around it.

When $S$ is differentiable and the relevant densities are positive, write $g_x=G_x'$ and $f_b=F_b'$. For $T=F_b^{-1}\circ G_x^{-1}\circ F_b$, the inverse satisfies $S=F_b^{-1}\circ G_x\circ F_b$, so
\begin{equation}
 S'(y)=\frac{g_x(F_b(y\mid x))f_b(y\mid x)}{f_b(S(y)\mid x)},
 \qquad c_T(y,y+h)=S'(y)|h|+o(|h|).
\end{equation}
The CDF representation remains valid without these derivative conditions. It shows why geometry can make some adverse movements more expensive and others less expensive.

\subsubsection{A loss-sensitive robust premium}

Hold the source scenarios and fitted map fixed, and let $\widehat\P^{(m)}=m^{-1}\sum_i\delta_{T(s_i)}$. Define the transported source ball $\mathcal U_T^{(m)}(\varepsilon)\coloneqq T_\#\{\Q_0:\W(\Q_0,\Pbm)\le\varepsilon\}$ and write $R_T(z,\varepsilon)\coloneqq R(z;\mathcal U_T^{(m)}(\varepsilon))$ for its robust upper risk. Then define
\begin{equation}
\label{eq:app-premium-coefficient}
 \kappa_T(z)=\max_{i\in[m]}\sup_{s\ne s_i}
 \frac{[\ell(z,T(s))-\ell(z,T(s_i))]_+}{|s-s_i|}.
\end{equation}
For continuous loss and transport, the robust premium satisfies
\begin{equation}
\label{eq:app-premium}
 R_T(z,\varepsilon)-\mathcal L(z;\widehat\P^{(m)})\le\kappa_T(z)\varepsilon,
 \qquad
 \lim_{\varepsilon\downarrow0}\frac{R_T(z,\varepsilon)-\mathcal L(z;\widehat\P^{(m)})}{\varepsilon}=\kappa_T(z),
\end{equation}
where the limit may be infinite. To prove the upper bound for finite $\kappa_T(z)$, disintegrate a feasible coupling over the empirical source atoms. Each displacement from $s_i$ increases the loss by at most $\kappa_T(z)|s-s_i|$, and integrating gives the bound. For the reverse limit, choose any loss-increasing pair $(s_i,s)$ and move mass $\theta=\varepsilon/|s-s_i|$ from $s_i$ to $s$. This is feasible whenever $\theta\le1/m$ and increases the risk by exactly its secant ratio times $\varepsilon$. Taking the supremum over such pairs proves the limit. If a pair attains a positive finite maximum, equality in the first bound holds for $0\le\varepsilon\le|s-s_i|/m$.

At the same corrected center, ordinary geometry replaces the denominator in \eqref{eq:app-premium-coefficient} by $|T(s)-T(s_i)|$. The comparison is therefore about the cost of loss-increasing movements. A small transportation budget can move a small amount of mass far away, which is why secant slopes, rather than only local derivatives, determine this premium.

\subsubsection{Strict improvement for procurement}

For $\ell(z,y)=c_pz+c_s(y-z)_+$ on $z,y\in[a,b]$, assume $0<c_p<c_s$. Let $z_0$ minimize the fixed-distribution risk $\mathcal L(z;\widehat\P^{(m)})$, and put $v_0=\mathcal L(z_0;\widehat\P^{(m)})$. Let $v_C(\varepsilon_C)$ be ordinary Wasserstein DRO at that center, and let $v_{\rm DT}(\varepsilon_T)$ use the image of the source ball. Define
\begin{equation}
\label{eq:gamma}
 \gamma_T(z_0)=\max_{i\in[m]}\sup_{s\ne s_i}
 \frac{[(T(s)-z_0)_+-(T(s_i)-z_0)_+]_+}{|s-s_i|}.
\end{equation}

\begin{proposition}[Geometry improvement at a common corrected center]
\label{prop:geometry-gain}
If
\begin{equation}
\label{eq:geometry-condition}
 \varepsilon_C<\frac{c_pb-v_0}{c_s},
 \qquad \gamma_T(z_0)\varepsilon_T<\varepsilon_C,
\end{equation}
then
\begin{equation}
\label{eq:geometry-gain}
 v_C(\varepsilon_C)-v_{\rm DT}(\varepsilon_T)
 \ge c_s[\varepsilon_C-\gamma_T(z_0)\varepsilon_T]>0.
\end{equation}
The comparison uses any radii for which the two ambiguity sets are valid on a common event. In particular, \Cref{app:containment} supplies the same radius for both sets, in which case $\gamma_T(z_0)<1$ suffices together with the first condition.
\end{proposition}

We first compute the ordinary-DRO benchmark. For any nominal law $\P_0$ on $[a,b]$, with nominal optimal value $v_0$, scalar Wasserstein duality gives
\begin{equation}
\label{eq:cap}
 \min_z R\left(z;\{\Q:\W(\Q,\P_0)\le\varepsilon\}\right)=\min\{v_0+c_s\varepsilon,c_pb\}.
\end{equation}
For a direct derivation, multipliers $\lambda>c_s$ cannot improve on $\lambda=c_s$, because the shortage loss is $c_s$-Lipschitz. Write $\lambda=c_sk$ with $k\in[0,1]$. Moving below a source outcome $S$ cannot improve the inner maximization, and convexity on $[S,b]$ reduces it to the endpoints. Its normalized value is
\begin{equation}
 \max\{(S-z)_+,(b-z)-k(b-S)\}=[b-z-k(b-S)]_+.
\end{equation}
For $k>0$, substitute $z=(1-k)b+kt$, obtaining the objective
\begin{equation}
 (1-k)c_pb+k\bigl[c_pt+c_s\E(S-t)_++c_s\varepsilon\bigr].
\end{equation}
The feasible interval for $t$ includes $[a,b]$; extending below $a$ cannot improve nominal procurement risk because its slope there is $c_p-c_s<0$. Minimizing in $t$ thus gives $(1-k)c_pb+k(v_0+c_s\varepsilon)$. At $k=0$ the minimum is $c_pb$. Minimizing over $k\in[0,1]$ proves \eqref{eq:cap}.

For DT-DRO, \eqref{eq:app-premium} applied at $z_0$ gives $v_{\rm DT}(\varepsilon_T)\le v_0+c_s\gamma_T(z_0)\varepsilon_T$. The first condition in \eqref{eq:geometry-condition} places ordinary DRO below its boundary-saturation threshold, so \eqref{eq:cap} gives $v_C(\varepsilon_C)=v_0+c_s\varepsilon_C$. Subtracting proves \eqref{eq:geometry-gain}. On a common coverage event, both objectives are valid upper bounds on the population risk of their own decisions. The result compares their robust certificates, not the realized losses of those different decisions.

A useful sufficient condition is that $T$ is absolutely continuous and $T'(s)\le\gamma<1$ almost everywhere on $\{s:T(s)>z_0\}$. Then $(T(s)-z_0)_+$ is $\gamma$-Lipschitz, so $\gamma_T(z_0)\le\gamma$. Expansion is allowed where shortage is zero. By contrast, an increasing homeomorphism of $[a,b]$ that is globally nonexpansive must be the identity: comparing $T(s)$ with each fixed endpoint yields both $T(s)\le s$ and $T(s)\ge s$. Thus global contraction is not the explanation for a nontrivial geometric benefit.

\paragraph{An exact illustration.}
Take $\ell(z,y)=z+20(y-z)_+$ on $[0,1]$, one source scenario $s_1=1/2$, and the increasing piecewise-linear map through $(0,0)$, $(1/2,4/5)$, $(3/4,19/20)$, and $(1,1)$. Both methods have center $\delta_{4/5}$ and radius $\varepsilon=3/200$. Ordinary DRO has value $1$ and selects $z=1$. For $z\in[4/5,1]$, DT-DRO has worst-case risk
\begin{equation}
 R_T(z,\varepsilon)=z+20\varepsilon\max\{4(19/20-z)_+,2(1-z)\}.
\end{equation}
The two adverse movements meet at $z=9/10$, giving the unique optimum $v_{\rm DT}=24/25<1$; decisions below $4/5$ have strictly larger objective. This calculation isolates geometry at a common center. It is not a claim of statistical coverage for the illustrative singleton source law.

\section{Experimental details}
\label{app:exp}

We describe the baselines, data-generating mechanisms, fitted models, radius-selection rules, and numerical solvers used in the experiments. The synthetic study considers four forms of predictive misspecification, while the outage study uses a representative weighted demonstration sample to reduce computation. We distinguish the diagnostic construction, its numerical implementation, and the empirical selection of ambiguity radii. The experimental radius-selection rules are separate from the bootstrap procedure in \Cref{app:radius}.

\subsection{Baseline definitions}
\label{app:baselines}

For a nominal law $\P_0$, write $\mathcal B(\P_0,\varepsilon)=\{\Q:\W(\Q,\P_0)\le\varepsilon\}$. At a given context, let $\Pbm$ be the empirical fixed-base law, let $\widetilde\P_{n,x}^{(m)}$ be the empirical refitted-base law, and let $\Phm=(\Th)_\#\Pbm$. Each radius-bearing method minimizes worst-case expected loss over its specified set, whereas DT-PTO minimizes expected loss at its single corrected center:
\begin{center}
\begin{tabular}{p{0.23\linewidth}p{0.69\linewidth}}
\toprule
Method & Distributional model used for optimization \\
\midrule
Fixed W-DRO & $\mathcal B(\Pbm,\varepsilon_{\rm fixed})$. The original fitted predictor remains unchanged. \\[1mm]
Retrained W-DRO & $\mathcal B(\widetilde\P_{n,x}^{(m)},\varepsilon_{\rm refit})$. The same predictive family is refitted on the historical observations available to that experiment. \\[1mm]
DT-PTO & $\{\Phm\}$. The diagnostic correction is retained, but the ambiguity radius is zero. \\[1mm]
DT-center W-DRO & $\mathcal B(\Phm,\varepsilon_{\rm center})$. This uses the same corrected center as DT-DRO, with ordinary outcome-space geometry. \\[1mm]
DT-DRO & $(\Th)_\#\mathcal B(\Pbm,\varepsilon_{\rm DT})$. The loss is evaluated at transported outcomes and the transportation budget in the source coordinates. \\
\bottomrule
\end{tabular}
\end{center}
The center-only and DT-DRO methods share the same corrected center, although their empirically selected radii need not coincide. The synthetic study additionally evaluates a \emph{Retrained + DT} method by fitting a new residual diagnostic to the refitted base; its results are retained but excluded from the displayed comparison. Experimental refitting uses the original training observations together with the available additional history, rather than discarding the original training sample.

\subsection{Four-setting synthetic experiment}
\label{app:synthetic_experiment}

\subsubsection{Conditional distributions and sampling}

All four synthetic settings have scalar outcomes in $[0,1]$ and independent contexts $X\sim\mathrm{Unif}(0,1)$. Decisions and density comparisons are evaluated at $x_0=0.55$. Write $d(x)=0.10(x-x_0)$, $\operatorname{TN}_{[0,1]}(\mu,\sigma^2)$ for a Gaussian distribution conditioned to lie in $[0,1]$, and $\operatorname{TG}_{[0,1]}(\mu,s)$ for an analogously truncated right-Gumbel distribution with location $\mu$ and scale $s$. We use the following settings, in the same order as the columns of \Cref{fig:synthetic}.

\paragraph{Gaussian location shift.}
The training and target populations differ in location:
\begin{equation}
\label{eq:app-synthetic-shift}
\begin{aligned}
 Y_{\rm train}\mid X=x&\sim\operatorname{TN}_{[0,1]}(0.42+d(x),0.10^2),\\
 Y_{\rm cal}\mid X=x&\sim\operatorname{TN}_{[0,1]}(0.57+d(x),0.10^2).
\end{aligned}
\end{equation}
This setting changes the population between base training and calibration without changing the underlying truncated-Gaussian family.

\paragraph{Heteroskedastic Gaussian misspecification.}
Training, calibration, and evaluation use the same conditional distribution,
\begin{equation}
\label{eq:app-synthetic-heteroskedastic}
 Y\mid X=x\sim\operatorname{TN}_{[0,1]}(0.43+d(x),\sigma(x)^2),
 \qquad
 \sigma(x)=0.055+0.09\exp\!\left[-\frac{(x-x_0)^2}{2(0.12)^2}\right].
\end{equation}
The conditional scale has a smooth local increase centered at $x_0$, where $\sigma(x_0)=0.145$. The Gaussian base model uses a single global scale, so refitting that model does not represent this context-dependent scale pattern.

\paragraph{Gumbel misspecification.}
Training and target outcomes follow
\begin{equation}
\label{eq:app-synthetic-gumbel}
 Y\mid X=x\sim\operatorname{TG}_{[0,1]}(0.38+d(x),0.105).
\end{equation}
Here the population is unchanged, but the Gaussian predictive family does not represent the asymmetric Gumbel shape.

\paragraph{Rare upper-tail mixture.}
Training and target outcomes follow
\begin{equation}
\label{eq:app-synthetic-mixture}
\begin{aligned}
 Y\mid X=x&\sim(1-\pi(x))\operatorname{TN}_{[0,1]}(0.33,0.075^2)
 +\pi(x)\operatorname{TN}_{[0,1]}(0.84,0.035^2),\\
 \pi(x)&=\min\{0.999,\max\{0.001,0.08+0.10(x-x_0)\}\}.
\end{aligned}
\end{equation}
At $x_0$, the upper component has probability $0.08$. A single Gaussian base cannot separately represent the main body and the rare upper component. All four distributions are bounded by construction; the Gumbel and mixture settings concern distributional shape and upper-tail misspecification within this support. Gaussian and Gumbel observations use inverse transforms of their truncated CDFs, while the mixture quantile is computed using 60 bisection steps on $[0,1]$.

\paragraph{Sampling and replication.}
Each independent replication uses 1,000 base-training observations, a nested calibration pool, and an independent validation sample of 500 target-population outcomes at $x_0$. We partition the validation sample in its generated order into $B_{\rm val}=50$ disjoint batches of $q=10$ outcomes for empirical coverage calibration. A separate sample of 500 independent outcomes, partitioned in the same way, evaluates batch-average coverage after radius selection and is never used for fitting or tuning. The calibration sample of size $n$ is a prefix of the same pool; the validation and coverage-test samples are reused across sample sizes within a replication. The configured experiment uses 100 replications and
\begin{equation}
 n\in\{100,168,283,476,800,1345,2263,3805,6400\}.
\end{equation}
The base seed is 20260924. Separate random streams generate the training, calibration, validation, and coverage-test data for each setting and replication. Within each replication and sample size, the competing methods share the calibration observations, validation and coverage-test batches, and uniform probabilities used to generate source scenarios. This pairing reduces variation unrelated to the method comparison. The default configuration runs all four settings afresh rather than only redrawing stored results.

\subsubsection{Base fitting and the local diagnostic}

The base predictor is a Gaussian GLM with an identity link, an intercept, a linear effect of $x$, and constant variance. It is fitted using at most 100 iterations and tolerance $10^{-8}$. The fixed model is trained once per scenario and replication. The refitted model uses the union of the base-training sample and the first $n$ calibration observations, with the preceding fit used as a warm start. Although fitting uses the Gaussian GLM, all CDF, density, and quantile evaluations in this experiment truncate and renormalize the predictive distribution to $[0,1]$.

The diagnostic starts from $U_j=F_b(Y_j\mid X_j)$. At $x_0$, the local estimator retains the
\begin{equation}
 k_n=\min\{n,\max\{30,\lceil n^{0.8}\rceil\}\}
\end{equation}
nearest calibration contexts in Euclidean distance, with equal normalized weights in this unweighted experiment. The experiment-specific transport routine maps their PIT values back to pseudo-outcomes $\widetilde Y_j=F_b^{-1}(U_j\mid x_0)$ and smooths those pseudo-outcomes on the bounded outcome scale. In particular, if $H_h(y;t)$ is the CDF of a $\mathcal N(t,h^2)$ kernel conditioned to $[0,1]$, the corrected CDF used to construct the map is
\begin{equation}
\label{eq:app-synthetic-corrected-cdf}
 \widehat F_{x_0}(y)=(1-\zeta_n)\frac1{k_n}\sum_{j\in I_n(x_0)}H_{h_n}(y;\widetilde Y_j)+\zeta_nF_b(y\mid x_0),
 \quad h_n=0.12k_n^{-1/5},\quad\zeta_n=\min\{0.05,(k_n+1)^{-1}\}.
\end{equation}
Thus the fitted correction is constructed from conditional PIT values but smoothed after mapping them back to the outcome scale, rather than by directly smoothing the PIT values on $[0,1]$.

A small numerical regularization is applied before constructing the transport. We evaluate \eqref{eq:app-synthetic-corrected-cdf} on 257 equally spaced outcome knots $y_j=j/256$, clip the probabilities to $[0,1]$, take their cumulative maximum, and pin the endpoints to zero and one. If $f_j$ denotes the resulting CDF values, we then form
\begin{equation}
\label{eq:app-synthetic-numerical-mixture}
 f_j^{\rm reg}=(1-\eta)f_j+\eta\frac{j}{256},
 \qquad \eta=10^{-8},\qquad j=0,\ldots,256.
\end{equation}
This perturbation prevents repeated source knots caused by probabilities rounding to zero or one. The source knots are $s_j=F_b^{-1}(f_j^{\rm reg}\mid x_0)$, with endpoints fixed at zero and one, and linear interpolation through $(s_j,y_j)$ defines $\widehat T_{x_0}$. Strict increase of the source knots is checked before optimization. The density panel displays the smooth KDE-plus-base density underlying \eqref{eq:app-synthetic-corrected-cdf}, before the $10^{-8}$ numerical perturbation, whereas optimization uses the piecewise-affine transport. The same fixed-base diagnostic is shared by DT-PTO, DT-center W-DRO, and DT-DRO. A separate diagnostic is fitted for the Retrained + DT ablation.

\subsubsection{Loss, radius choice, and robust solver}

All synthetic methods use $\ell(z,y)=z+20(y-z)_+$ with $z,y\in[0,1]$. The population-optimal decision is the true conditional $0.95$ quantile. The center is represented by $m=\max\{40,n\}=n$ source scenarios for the displayed sample-size ranges. Shared independent uniform probabilities are transformed through the corresponding fixed or refitted truncated-Gaussian quantile functions.

We select each radius-bearing method's radius by empirical certificate coverage rather than by minimizing validation loss. For method $j$ and candidate radius $\varepsilon$, let $\widehat z_j(\varepsilon)$ and $\widehat v_j(\varepsilon)$ be its optimized decision and robust objective. A validation batch is covered when its mean realized loss does not exceed that objective. With validation outcomes $Y^{\rm val}_{br}$, define
\begin{equation}
\label{eq:app-synthetic-coverage}
 \widehat C_j^{\rm val}(\varepsilon)
 =\frac{1}{B_{\rm val}}\sum_{b=1}^{B_{\rm val}}
 \1\left\{\frac1q\sum_{r=1}^q\ell\bigl(\widehat z_j(\varepsilon),Y^{\rm val}_{br}\bigr)
 \le\widehat v_j(\varepsilon)\right\},
 \qquad B_{\rm val}=50,\quad q=10.
\end{equation}
The same fixed batches are used for every method and radius. This mirrors the group-average loss criterion in the outage study, with independent batches at a fixed context replacing substation-month groups. The numerical comparison permits an additive tolerance of $10^{-9}$ on the robust objective.

The initial common radius grid is
\begin{equation}
\label{eq:app-synthetic-radius-grid}
 \mathcal E_{n,m}^{(0)}
 =\left\{\min\{a s_{n,m},1\}:a\in\{0,0.02,0.05,0.10,0.20,0.50,1.00\}\right\},
 \quad s_{n,m}=\sqrt{\frac{\log(k_n+1)}{k_n}}+\sqrt{\frac{\log(m+1)}m}.
\end{equation}
We test distinct candidates in increasing order. If none attains the target, we repeatedly double the largest tested radius, capping it at the support diameter $1$, until the target is met. Let $\mathcal E_{n,m}$ denote this expanded grid. The selected radius is
\begin{equation}
\label{eq:app-synthetic-coverage-radius}
 \widehat\varepsilon_j
 =\min\left\{\varepsilon\in\mathcal E_{n,m}:\widehat C_j^{\rm val}(\varepsilon)\ge0.90\right\}.
\end{equation}
The decision is reoptimized at every candidate, so the search does not assume that empirical coverage is monotone in the radius. Extending the grid prevents persistent model error from being excluded by a maximum candidate radius that vanishes with $n$. At radius $1$, any distribution on $[0,1]$ is reachable from the source center; the fully protective decision $z=1$ attains coverage one for this procurement loss. Mean validation loss is retained only as a descriptive statistic and is not used to select among candidates. DT-PTO retains radius zero and is not calibrated to the coverage target.

The rule attains at least $90\%$ coverage on the selected validation batches, not necessarily exactly $90\%$, because both the candidate radii and empirical coverage are discrete. We separately evaluate the selected decision and objective on the independent coverage-test batches and record whether $\widehat v_j\ge\mathcal L^\star(\widehat z_j)$. Neither quantity enters radius selection. Coverage of a finite-batch average is distinct from coverage of the population mean and from containment of the entire conditional distribution. Thus this empirical rule does not replace the diagnostic-confidence assumptions of \Cref{thm:radius,thm:value-certificate}.

The synthetic solver exploits the same endpoint reduction as \Cref{cor:lp-reduction}. It further reduces the procurement objective to a scalar profile. For source samples $\nu_i$ and an increasing piecewise-affine $T$, write $\theta=\lambda/c_s$ and define
\begin{equation}
\label{eq:app-adjusted-outcomes}
 w_i(\theta)=\max\left\{T(\nu_i),\ \max_{\tau_j\ge\nu_i}\bigl[T(\tau_j)-\theta(\tau_j-\nu_i)\bigr]\right\}.
\end{equation}
For $s<\nu_i$, monotonicity of $T$ and of the shortage loss makes movement below $\nu_i$ unhelpful. For $s\ge\nu_i$, maximizing the two affine pieces of the hinge gives
\begin{equation}
 \sup_s\{c_pz+c_s(T(s)-z)_+-\lambda|s-\nu_i|\}
 =c_pz+c_s(w_i(\theta)-z)_+.
\end{equation}
The $w_i(\theta)$ are transportation-penalized outcome values, not necessarily realizations of an adversarial distribution. Conditional on $\theta$, minimizing the empirical hinge loss chooses the empirical critical quantile of these values, at probability $1-c_p/c_s=0.95$. The remaining objective is minimized over $0\le\theta\le\max_j(T(\tau_{j+1})-T(\tau_j))/(\tau_{j+1}-\tau_j)$. The implementation uses bounded scalar minimization with absolute tolerance $10^{-9}$ and at most 200 iterations, and compares the result with both endpoints. A suffix maximum over adjusted knots computes all the inner maxima efficiently. For an identity transport, the solver compares the nominal empirical value plus $c_s\varepsilon$ with the boundary-procurement value $c_p$ on $[0,1]$; it retains a nominal optimizer in the former case and chooses $z=1$ in the latter. Zero radius uses the nominal empirical quantile.

The inner continuum is therefore eliminated exactly for the fitted piecewise-affine map; the outer scalar minimization is numerical. This does not remove estimation error in the map or the empirical-center approximation. It is also different from assuming that arbitrary worst-case scenarios can be selected once and then held fixed while optimizing the decision.

\subsubsection{Evaluation and uncertainty bands}

The reported synthetic regret is
\begin{equation}
 \mathcal L^\star(z)-\min_{z'\in[0,1]}\mathcal L^\star(z').
\end{equation}
Expected losses are evaluated by averaging over 20,000 midpoint quantiles of the known true conditional law at $x_0$. This numerical integration is separate from the $m$ scenarios used by each decision method. Plotted means average the independent replication regrets; the bands are pointwise $95\%$ Student-$t$ intervals $\overline{\mathrm{Reg}}\pm t_{0.975,B-1}s_{\mathrm{Reg}}/\sqrt B$, where $B=100$ is the replication count. The plotted lower endpoint is truncated at zero. Density estimates are averaged across replications at the largest calibration size and evaluated on a 601-point outcome grid.

We retain replication-level decisions, radii, robust objectives, true and oracle costs, diagnostic neighborhood sizes, transport slopes, and density evaluations. We additionally record validation and independent-test batch coverage, population-risk certificate indicators, radius-search expansion, and boundary decisions, and summarize these quantities across replications. These records distinguish decision performance, achieved empirical protection, numerical approximation, and the selected degree of robustness.

\subsection{Real-data experiment: the weighted demonstration configuration}
\label{app:outage_experiment}

\subsubsection{Data preparation and the demonstration sample}

The data comprise outage-event records, monthly substation covariates, and network topology. Repeated records are collapsed using event time, substation, location, and duration, with customer interruptions summed, giving 108,188 distinct physical events before covariate matching. The response is duration in hours capped at 48, $Y=\min\{\text{duration in minutes}/60,48\}$. The cleaned full modeling table contains 62,890 matched observations over February 2014 through August 2024. The demonstration uses a weighted subset of 6,624 observations, assembled to retain tail behavior while reducing runtime.

The demonstration sample first selects six representative substations. Candidate substations must appear in at least eight distinct years; all matched substations are used as candidates if too few qualify. We standardize summaries of log event count, median, 90th- and 99th-percentile durations, severe-event frequency, customer impact, and location, apply $K$-means with six clusters and 30 initializations, and select the member nearest each cluster center. Each representative receives its cluster's event count divided by its own event count as a representation weight. Within each representative substation-month, all events at or above the full modeling table's $95\%$ duration quantile, approximately $8.01$ hours, are retained. Ordinary events are sampled without replacement, targeting 25 total observations per month while retaining at least five ordinary observations when available. Kept tail events have within-month weight one; sampled ordinary events receive the inverse of their sampling fraction. The analysis weight is the product of representation and within-month weights.

This cohort is assembled retrospectively from the available data; subsequent model fitting and decisions use only the prescribed past observations within that cohort. The tail-retention cutoff used to build the sample is distinct from the static-training threshold used for severe-outage evaluation. The resulting splits contain 1,578 demonstration observations in the 2014--2016 static training sample and 5,576 in the rolling pool beginning in 2016, with 530 observations overlapping during 2016. Monthly evaluation from January 2017 through August 2024 uses 5,046 demonstration observations across 92 months. These overlapping training and rolling totals should not be added as disjoint counts.

\subsubsection{Predictive covariates and the Gamma base model}

The numerical covariates are one-month-lagged precipitation, maximum and minimum temperature, and wind speed; tree coverage; $\log(1+\text{pole count})$; $\log(1+\text{operations-and-maintenance spending})$; and sine and cosine seasonal terms. Substation indicators are added. Numerical means and scales are estimated using weighted static-training observations and held fixed. The implementation retains a full-rank subset of the encoded columns, including an intercept, because some fixed substation attributes are collinear with the indicators. Contemporaneous outage-summary measures are excluded from the predictors.

The base is a Gamma GLM with a log link. Writing $\mu(x)=\exp(\beta_0+x^\top\beta)$ and $\phi$ for fitted dispersion, the distribution used by the implementation is
\begin{equation}
 Y\mid X=x\sim\operatorname{Gamma}(k,\vartheta(x)),
 \qquad k=1/\phi,\quad\vartheta(x)=\phi\mu(x),
 \qquad\operatorname{Var}(Y\mid X=x)=\phi\mu(x)^2.
\end{equation}
The model is fitted using the analysis weights as frequency weights, at most 150 iterations, and tolerance $10^{-8}$. The fixed model uses the static 2014--2016 sample. At each evaluation month, the refitted model uses static observations before January 2016 together with rolling observations from 2016 onward strictly before that month, avoiding double counting the overlap. Fits are warm-started from the preceding month.

The Gamma CDF and quantile function are evaluated without truncation. Generated source values are subsequently clipped to $[10^{-6},48]$, and transported values used by the solver are clipped to $[0,48]$. This numerical convention differs from the truncated-and-renormalized Gaussian predictions in the synthetic experiment.

\subsubsection{Monthly conditional PIT estimation}

At each month, the fixed-base PIT values from the available rolling history are paired with their contexts to fit a histogram gradient-boosting classifier. Each observation is expanded at eight independent uniform probability thresholds, together with the endpoints zero and one, and its analysis weight is repeated for each threshold. The binary response at threshold $u$ is $\1\{F_b(Y_i\mid X_i)\le u\}$; the endpoint responses are pinned to zero and one. The threshold coordinate is constrained to have a nonnegative monotonic effect. We use learning rate $0.065$, at most 40 boosting iterations, at most 15 leaves, minimum leaf size 20, and $\ell_2$ regularization parameter one.

The fitted probabilities are evaluated on 65 equally spaced PIT thresholds, projected by isotonic regression, and pinned at the endpoints. Linear interpolation gives CDF evaluations, and a generalized inverse of this curve gives quantile evaluations. The transport is formed by composing this generalized inverse with the Gamma CDF and quantile function, with probability clipping at $10^{-10}$ and $1-10^{-10}$. The implementation enforces nondecrease but does not add a strict identity mixture for this boosting fit; flat portions and outcome clipping can therefore prevent the resulting numerical map from being a homeomorphism. The finite-grid calculations below are described as the implemented approximation, rather than as an automatic application of every invertible-map theorem.

A corresponding diagnostic is also fitted to the refitted Gamma model and used in its radius calculation. The five plotted methods do not include a Retrained + DT policy. The fixed-base diagnostic fit and the 2014--2016 fixed-base training data overlap in 2016; thus the real-data protocol is not an independent base-training/calibration experiment of the form assumed in \Cref{ass:diagnostic}.

\subsubsection{Decisions and the finite-grid solver}

A single decision is chosen for each observed substation-month and evaluated on all outages at that context in the month. The loss is $\ell(z,y)=0.10z+(y-z)_+$, with $z\in[0,48]$, targeting the nominal $0.90$ quantile. At each context, the base distribution is represented using 24 midpoint probabilities $(j-1/2)/24$, transformed by its quantile function and then clipped as described above. DT-PTO uses the upper order-statistic convention for the empirical $0.90$ quantile of the corrected center; for 24 scenarios, this selects its 22nd ordered outcome.

The robust solver uses a finite source grid consisting of 65 equally spaced points in $[10^{-6},48]$ augmented by the source scenarios. For DT-DRO, its adversarial outcomes are the clipped transported grid points and its distance matrix is $|s_k-\nu_j|$. Ordinary W-DRO instead measures distances in outcome coordinates using a grid on $[0,48]$ augmented by the nominal scenarios; the center-only method additionally includes transported grid points. At each candidate decision, the solver minimizes the scalar multiplier problem in
\begin{equation}
\label{eq:app-real-grid}
 \min_{z\in\mathcal Z_{\rm grid},\lambda\ge0}
 \left\{\lambda\varepsilon+\frac1m\sum_{j=1}^m\max_k\bigl[\ell(z,y_k)-\lambda d_{kj}\bigr]\right\},
\end{equation}
where $\mathcal Z_{\rm grid}$ contains the 65-point decision grid, nominal outcomes, and adversarial outcomes. A bound derived from positive loss-to-distance ratios brackets the multiplier; 16 bisection steps are used, based on whether the active average transportation cost exceeds the radius. The smallest resulting objective among candidate decisions is selected.

Unlike the synthetic endpoint calculation, this procedure discretizes both the adversarial search and the decision search. It is not claimed to be the exact continuum LP in \Cref{cor:lp-reduction}. A finite-grid robust objective need not remain an upper bound for the continuous-outcome problem, which is one reason to distinguish its empirical certificate evaluation from the analytical guarantee in \Cref{thm:value-certificate}.

\subsubsection{Initial radii and past-only multiplier calibration}

The real-data radius is an empirical tuning rule. Let $N_t$ be the sum of historical analysis weights and let $s_x$ be the interquantile range between probabilities $0.10$ and $0.90$ of the source-scenario vector, floored at $0.25$. To summarize the fitted diagnostic, evaluate corrected PITs $p_i=\widehat G_{X_i}(U_i)$ on at most 1,500 approximately equally spaced historical rows. Sort them as $p_{(1)}\le\cdots\le p_{(N)}$, carrying their weights $w_{(i)}$, and define
\begin{equation}
 E_t=\frac{\sum_{i=1}^Nw_{(i)}|p_{(i)}-r_i|}{\sum_{i=1}^Nw_{(i)}},
 \qquad r_i=\frac{\sum_{j\le i}w_{(j)}-w_{(i)}/2}{\sum_{j=1}^Nw_{(j)}}.
\end{equation}
This is an in-sample, aggregated corrected-PIT discrepancy, not a confidence bound for $\|\widehat G_x-G_x^\star\|_\infty$ at a fixed context. The initial radii are
\begin{equation}
\label{eq:app-real-initial-radii}
 \varepsilon^{\rm DT}_{x,t}=\max\left\{0.025,\frac{0.04s_x}{\sqrt{\max(N_t,1)/1000}}+s_xE_t\right\},
 \qquad
 \varepsilon^{\rm ordinary}_{x,t}=\varepsilon^{\rm DT}_{x,t}+\frac1m\sum_{j=1}^m|\nu_j-\widehat T_x(\nu_j)|,
\end{equation}
with transported values clipped to $[0,48]$. DT-DRO and DT-center W-DRO use the first baseline radius; Fixed W-DRO uses the second. Retrained W-DRO recomputes the second expression using its own refitted model and diagnostic. Here $N_t$ is a weighted count used in the heuristic rate term, not a claim about an independent effective sample size.

Each radius-bearing method maintains its own multiplier. A substation-month is covered when its analysis-weighted mean realized loss does not exceed the objective computed before observing that month. Let $C_{j,t}$ be the fraction of covered substation-months for method $j$ over the preceding 12 months, with each context contributing equally to that fraction. With target $C^\star=0.90$, the multiplier update is
\begin{equation}
 a_{j,t}=\operatorname{clip}_{[0,8]}\!\left[
 a_{j,t-1}+\frac{C^\star-C_{j,t}}
 {(1-C^\star)\1\{C_{j,t}\le C^\star\}+C^\star\1\{C_{j,t}>C^\star\}}
 \right].
\end{equation}
Before three validation months are available, the multiplier is one. The radius entering \eqref{eq:app-real-grid} is the baseline radius times $a_{j,t}$. All multiplier updates use past evaluation months only. Targeting $90\%$ in this feedback rule is not a guarantee of exactly $90\%$ realized coverage, and this context-average loss event differs from distributional containment.

\subsubsection{Regret and uncertainty summaries}

The perfect-information decision for an observed duration $y$ is $z^{\rm PI}=y$, with loss $0.10y$. Event-level perfect-information regret is therefore $\ell(z,y)-0.10y$, clipped below at zero for numerical safety. For each month and method, the displayed statistic is the analysis-weighted mean over that month and the preceding 11 months. The severe panel restricts the same calculation to durations exceeding the weighted $95\%$ quantile of the static training sample; the weighted demonstration sample gives a threshold of approximately $6.67$ hours. This tail subset is not a CVaR optimization objective.

The pointwise bands use 300 substation-cluster bootstrap replications. For each rolling window, we aggregate the weighted regret numerator and weight denominator within substations, resample those clusters with replacement, and recompute their ratio. The $2.5\%$ and $97.5\%$ empirical quantiles give the band. The base seed is 20260923. These are uncertainty summaries for the empirical regret comparison, not confidence bands for the conditional predictive law. The separate annual P--P diagnostics for 2018, 2020, 2022, and 2024 fit through the end of each displayed year and evaluate on that same year; they are in-sample diagnostics and are not used as prospective decision-performance evidence.

\subsection{Construction of the schematic in \texorpdfstring{\Cref{fig:transport}}{Figure 1}}
\label{app:distribution_space_schematic}

The illustration uses a two-dimensional coordinate field to display how a geometric adjustment can change the adverse distributions considered for a fixed decision. It does not simulate the statistical estimator or solve \eqref{eq:center-ot}. A point $q=(q_1,q_2)$ is an abstract display coordinate, not a verified isometric embedding of a probability distribution. The original set is a disk with radius $r_0=1.05$ and center $c_b=(-0.12,-0.08)$. Put $\theta_0=25^\circ$ and $c_D=c_b+0.90(\cos\theta_0,\sin\theta_0)$. The zero-potential location $q^\star=c_D$ is designated as the ground-truth marker. For $q-c_D=(r_1,r_2)$, define
\begin{align}
 u&=\cos\theta_0\,r_1+\sin\theta_0\,r_2,
 \qquad v=-\sin\theta_0\,r_1+\cos\theta_0\,r_2,\notag\\
 \widetilde u&=u+0.13[\sin(2.40v+0.55)-\sin(0.55)]\notag\\
 &\quad+0.06[\sin(3.10u-1.20v-0.35)-\sin(-0.35)],\notag\\
 \widetilde v&=v-0.30\sin(1.55u)
 +0.08[\sin(3.70u+1.40v+0.40)-\sin(0.40)].\notag
\end{align}
The scalar potential and the displayed diagnostic discrepancy are
\begin{align}
 \Psi(q)&=\tfrac12(0.52\widetilde u^2+1.45\widetilde v^2)
 [1+0.16\sin(1.90u+1.30v)+0.10\sin(3.30u-2.10v)]\notag\\
 &\quad+0.07(\widetilde u\widetilde v)^2
 +0.018[\sin(2.80u)\sin(2.20v)]^2,\notag\\
 \kappa(q)&=1-\exp[-\Psi(q)/0.55].
\end{align}
The coordinate map is $S(q)=q-0.16\nabla\Psi(q)$, with the gradient evaluated by centered differences of step $10^{-4}$. Applying $S$ to 900 equally spaced angles on the original circle produces the plotted deformed boundary. The image of the original center is checked to lie inside that boundary. The heatmap uses a $420\times420$ grid over $[-1.55,1.55]^2$, while the gray arrows sample the same displacement field on a $10\times10$ grid over $[-1.35,1.35]^2$, with their lengths rescaled for display. This is a single coordinate-map step, not an iterative optimization algorithm.

At the two selected locations $c_D+(0.12,0.25)$ and $(1.20,-1.10)$, the displayed discrepancies are approximately $0.0651$ and $0.9387$. Each inset draws $G_q(p)=p^{1+2.2\kappa(q)}$ at 300 equally spaced probabilities. These P--P curves are schematic and are not fitted from calibration samples. Their shaded areas satisfy
\begin{equation}
 \int_0^1|G_q(p)-p|\,\dd p=\frac12-\frac1{2+2.2\kappa(q)},
\end{equation}
which is monotone in the displayed discrepancy but not equal to it.

\paragraph{Adverse points and the displayed loss gaps.}
To obtain the two adverse-point annotations, fix $z_{\rm disp}=(0.6,1.0)^\top/\sqrt{1.36}$ and define a linear display loss
\begin{equation}
\label{eq:app-schematic-loss}
 J_{\rm disp}(q)=2.5-z_{\rm disp}^\top q,
 \qquad
 g_{\rm disp}(q)=J_{\rm disp}(q)-J_{\rm disp}(q^\star).
\end{equation}
The maximum of $J_{\rm disp}$ over the original disk is attained at $q_b^{\rm adv}=c_b-r_0z_{\rm disp}$. For the transported boundary, write $b(\theta)=c_b+r_0(\cos\theta,\sin\theta)$ and search for a maximizer of $J_{\rm disp}(S(b(\theta)))$ over $\theta\in[0,2\pi)$. We first evaluate 4,096 equally spaced angles and identify all grid-local maxima. Each candidate is refined within its neighboring angular interval by up to 60 golden-section iterations, stopping once the interval is narrower than $10^{-10}$. The largest refined value determines $q_T^{\rm adv}$. The resulting coordinates and display gaps are
\begin{equation}
\label{eq:app-schematic-regrets}
\begin{aligned}
 q_b^{\rm adv}&\approx(-0.660221,-0.980368),
 &g_{\rm disp}(q_b^{\rm adv})&\approx1.795815,\\
 q_T^{\rm adv}&\approx(-0.913179,-0.354389),
 &g_{\rm disp}(q_T^{\rm adv})&\approx1.389189.
\end{aligned}
\end{equation}
These are the values rounded to $1.80$ and $1.39$ in \Cref{fig:transport}. They compare the worst-case display loss with its value at the designated truth coordinate for the same fixed decision. They are not the population decision regret used in the synthetic experiment or the perfect-information regret used in the outage study. No predictive distribution is fitted, and no decision is reoptimized in this illustration. The coordinate field illustrates geometric adjustment; it is not established as the distribution-level action of the common outcome transport in \eqref{eq:udt}.

\end{document}